\pdfoutput=1

\documentclass[11pt]{article}

\usepackage{emnlp2021}

\usepackage{times}
\usepackage{latexsym}
\usepackage{graphicx}
\usepackage{tikz}
\usepackage{ctable}
\usepackage{float}
\usepackage{enumitem}
\usepackage{chngcntr}
\setitemize{noitemsep,topsep=0.1pt,parsep=0.1pt,partopsep=0.1pt}

\usepackage[T1]{fontenc}

\usepackage[utf8]{inputenc}

\usepackage{microtype}

\newcommand\matt[1]{\textcolor{purple}{}}

\definecolor{darkgreen}{rgb}{0.0, 0.5, 0.0}
\definecolor{byzantine}{rgb}{0.74, 0.2, 0.64}

\title{Leveraging Pretrained Models for Automatic Summarization of Doctor-Patient Conversations}

\author{
\textbf{Longxiang Zhang}$^1$ \quad \textbf{Renato Negrinho}$^2$ \quad 
\textbf{Arindam Ghosh}$^1$ \quad \textbf{Vasudevan Jagannathan}$^1$ \\ 
\textbf{Hamid Reza Hassanzadeh}$^1$ \quad \textbf{Thomas Schaaf}$^1$ \quad 
\textbf{Matthew R. Gormley}$^2$ 
\\ \\
\textrm{$^1$3M Health Information Systems} \quad \textrm{$^2$Carnegie Mellon University}
\\
\small \texttt{\{lzhang28,aghosh4,juggy,hhassanzadeh,tschaaf\}@mmm.com,  \{negrinho,mgormley\}@cs.cmu.edu}
}

\begin{document}
\maketitle

\begin{abstract}
  Fine-tuning pretrained models for automatically summarizing doctor-patient conversation transcripts presents many challenges: limited training data, significant domain shift, long and noisy transcripts, and high target summary variability. In this paper, we explore the feasibility of using pretrained transformer models for automatically summarizing doctor-patient conversations directly from transcripts. We show that fluent and adequate summaries can be generated with limited training data by fine-tuning BART on a specially constructed dataset. The resulting models greatly surpass the performance of an average human annotator and the quality of previous published work for the task.
  We evaluate multiple methods for handling long conversations, comparing them to the obvious baseline of truncating the conversation to fit the pretrained model length limit. We introduce a multistage approach that tackles the task by learning two fine-tuned models: one for summarizing conversation chunks into partial summaries, followed by one for rewriting the collection of partial summaries into a complete summary\footnote{Code is available at \url{https://github.com/negrinho/medical_conversation_summarization}}. Using a carefully chosen fine-tuning dataset, this method is shown to be effective at handling longer conversations, improving the quality of generated summaries. We conduct both an automatic evaluation (through ROUGE and two concept-based metrics focusing on medical findings) and a human evaluation (through qualitative examples from literature, assessing hallucination, generalization, fluency, and general quality of the generated summaries).
\end{abstract}


\section{Introduction}
\label{sec:introduction}

In recent years, pretrained transformer models \cite{lewis2019bart,devlin2018bert,zaheer2020big,gpt-3}  have been responsible for many breakthroughs in natural language processing (NLP) such as improved state-of-the-art performances for a broad range of tasks 
and the ability of training effective models for low-resource tasks. 
The demonstrated capability of transfer learning using large pretrained transformer models has led to widespread interest in leveraging these models in less standard NLP domains. 
Medical domains provide unique challenges and great potential for practical applications (e.g., see \citet{amin2020exploring} and \citet{huang2019clinicalbert}). 
Automatic generation of medical summaries from doctor-patient conversation transcripts presents several challenges such as the limited availability of supervised data, the substantial domain shift from the text typically used in pretraining, and potentially the long dialogues that exceed the length limitation of conventional transformers. 
Additionally, the model must have both extractive (e.g., such medications being taken, medication dosage, and numeric values of test results) and abstractive (e.g., the ability to determine the onset of a symptom from multiple conversation turns) capabilities.

Existing work on summarization from medical dialogue transcripts has achieved only limited success, both with pretrained models and otherwise. \citet{krishna2020generating} relied on extra supervision to train a classifier to extract noteworthy utterances that are relevant to the target summary and do not handle the long conversations with their pretrained models, and their example results suffer from inferior fluency. Other existing work relying extractive methods is poorly adjusted to the informal nature of dialogue and the fact that information might not be present in any single span from the conversation transcript. Due to this, it has not yet been established that pretrained models are able to successfully perform automatic summarization from doctor-patient conversation transcripts.

In this paper, we attempt to tackle the task of medical dialogue summarization by leveraging pretrained transformer models. We show that BART \cite{lewis2019bart} can be fine-tuned to generate highly fluent summaries of surprisingly good quality even with a small dataset of no more than 1000 doctor-patient conversations (Section~\ref{sec:dataset}). We overcome the input length limitations through a multistage fine-tuning approach in which the task of dialogue summarization is achieved in two steps: summarizing portions of input conversation and rewriting aggregated summaries of each portion (Section~\ref{sec:methods}). 
Our approach is simple as it amounts to fine-tuning pretrained model on appropriately constructed datasets. Despite its simplicity, it is effective at improving performance according to both automatic evaluation and human inspection (Section~\ref{ssec:model_comparison}-\ref{ssec:human_evaluation}) when compared to the baseline approach of simply truncating the input. 
We also observe good generalization of our fine-tuned models across medical domains and conversation lengths, as shown by example conversations from other papers tackling the same task such as \citet{krishna2020generating} and \citet{joshi2020dr} (Section~\ref{ssec:ood_example}). These examples also show the superior quality of our generated summaries.


\section{Dataset}
\label{sec:dataset}

The dataset used in this paper is based on a collection of more than 80000 de-identified doctor-patient conversations (both audio and transcript). 1342 conversations of two major specialties: internal medicine and primary care are annotated by medical scribes using our annotation environment specifically designed for the task. The scribes listen to the conversation audio and fill in necessary information in a simulated Electronic Health Record (EHR) system. The EHR simulator consists of 14 distinct sections such as History of Present Illness (HPI) and Review of System (ROS). 

We collect multiple references for each conversation, for a total of 21588 annotations. The dataset is split by conversation into train, development, and test with 939(15043), 201(3095), and 202(3450), respectively, where the values in parentheses are the number of HPI summaries in that split. Additional statistics are included in Appendix~\ref{ssec:data_stats}. 


We choose to use only the HPI section as our training target due to several observations: first, non-HPI sections are much less frequently filled by scribes, e.g., no more than 5\% of all annotations have covered ROS section; second, scribes are required to  write coherent paragraphs in the HPI section, whereas other sections might be structured as forms with most items being multiple choice; third, scribes are trained to cover non-HPI aspects like medication or physical examination in the HPI section if they are relevant to the "present illness" of the patient, making HPI section a good candidate for capturing most important medical findings in the conversation. 

Each conversation in our dataset has on average 15 reference HPI summaries from different scribes. One running example of a long conversation (with more than 2200 words) and three corresponding references are showcased in Appendix~\ref{ssec:running_example}. 
As can be seen from the example, different references can exhibit large variance in length and quality. For consistency, we select target reference summaries in the training set as follows: first, we leverage our rule-based system to extract medical findings from all reference summaries; then we select the reference with the most findings as target. 
While filtering out low-quality training summaries is expected to impact the performance of the fine-tuned models, we leave such study for future work.


\section{Methods}
\label{sec:methods}

The methods that follow can be broken down into single-stage and multistage. All models rely on fine-tuning of pretrained BART models, the difference being how the datasets used for fine-tuning are constructed. For the single-stage approach, conversation transcripts are serialized with doctor and patient roles annotated (i.e., the encoder consumes a single sequence for the conversation) and mapped directly to the target summary. Conversations longer than the transformer model length limit are simply truncated, leading to unrecoverable information loss. 
Despite the simplicity of this approach, it works remarkably well and it serves as a strong baseline to beat.
For the multistage approach, the conversation is first broken down into parts that are summarized independently by one model, and the resulting partial summaries are then aggregated and summarized into a final summary by another model. The methods that we propose in this class differ in how they break down the conversation into parts and therefore, the datasets that are used for fine-tuning their first stage model.

The multistage approach is motivated by the necessity of getting around the limited length budget of pretrained models along with the belief that medical findings covered in a summary is often present locally in a contiguous set of turns between the doctor and the patient, allowing each part to be summarized independently, with a later aggregation stage of all part summaries.


\subsection{Multistage summarization}
\label{ssec:multi_stage}

\begin{figure*}[tbhp]
    \centering
    \includegraphics[width=0.9\textwidth]{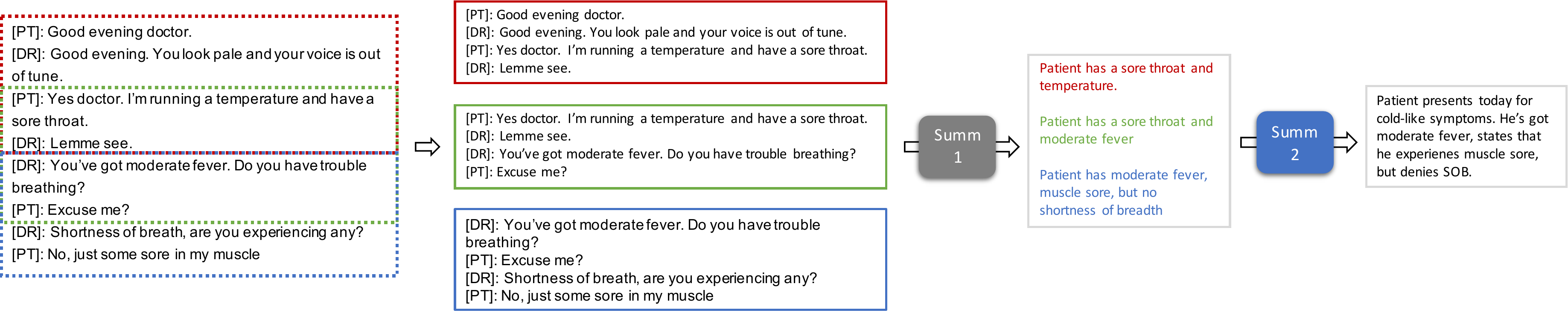}
    \caption{Multistage inference with SentBERT method. \textbf{Summ} stands for summarizer. The training target for \textbf{Summ 1} is a single sentence from the HPI summary. Complete summaries are used as target for \textbf{Summ 2} only.}
    \label{fig:illus_sentbert}
\end{figure*}

\begin{figure*}[tbhp]
    \centering
    \includegraphics[width=0.9\textwidth]{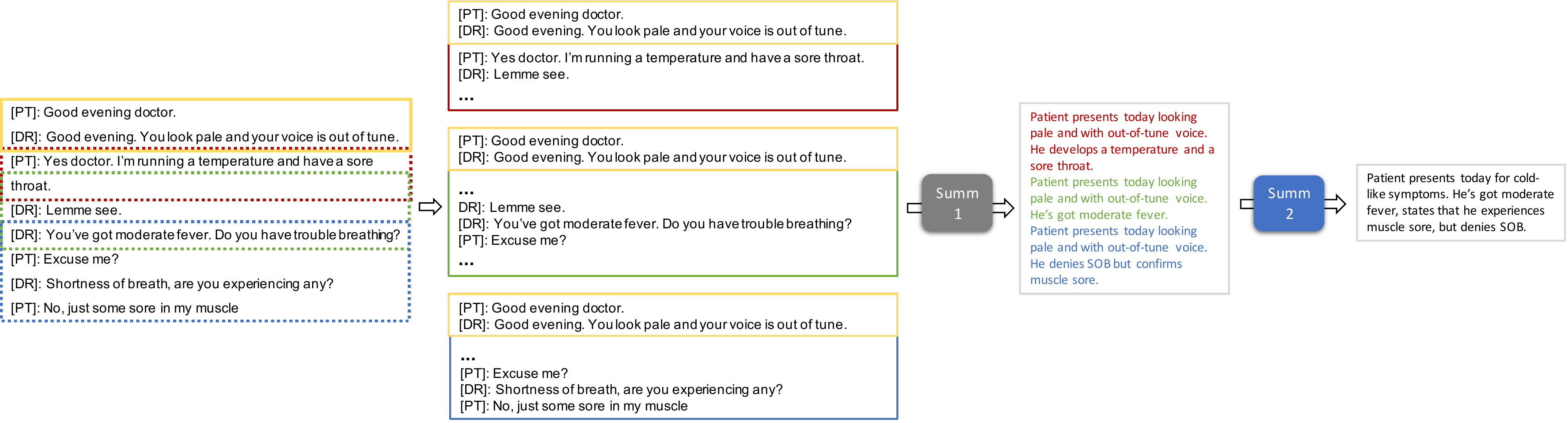}
    \caption{Multistage inference with Chunking method. \textbf{Summ} stands for summarizer. The same header (denoted by the yellow box) is added to the beginning of every chunk, serving as context, 
    and the complete summaries are used as targets for fine-tuning both \textbf{Summ 1} and \textbf{Summ 2}.}
    \label{fig:illus_chunking}
\end{figure*}



We experiment with two methods of breaking down the conversations into parts and setting up datasets for fine-tuning the first stage summarizer:

\paragraph{\textbf{SentBERT}.} We break all reference HPI summaries into individual sentences using the standard sentence splitter from the NLTK library \cite{nltk} and then create a collection of snippets of eight consecutive turns by sliding window over the conversation with stride one. Cosine similarity between each summary sentence and all the snippets is then calculated using their respective hidden representations generated by the pretrained Sentence-BERT model \cite{reimers2019sentence}. All snippets that have a similarity of 0.7\footnote{0.7 as the similarity threshold leads to most reasonable snippets by sample inspection.} or higher are then coalesced in case of overlap, and the longest such snippet is matched to the summary sentence. 
99.6\% of snippets generated in this way are within the input length limit, with an average of 230 tokens. One disadvantage of this method is that at inference time, we do not have reference summary to identify "similar" snippets. Therefore, the input to the second summarizer is created by first breaking each conversation into a set of 8-turn snippets with four turn overlap, and then generating single-sentence summaries from these snippets, and finally concatenating all generated sentences into a single paragraph\footnote{No additional post-processing steps are taken to filter out "noisy" sentences from potentially irrelevant snippets, we hypothesize that training in the second stage should instruct the summarizer on how to filter out those sentences automatically}. See Figure~\ref{fig:illus_sentbert} for an illustration on the inference procedure and Figure~\ref{fig:oak_sentbert} for examples of sentences generated for snippets.
%

%

\paragraph{\textbf{Chunking}.} 
We create chunks of transcript from each conversation where each chunk consists of two components: a fixed-length "header" that is selected from the beginning of the conversation, and is present in all chunks; a variable "body" that is created by a sliding scan of the rest of the conversation. A special ellipsis token "..." is added between any header and body that are not contiguous and at the end of every non-terminating chunk, marking the existence of transcript text that is not present in the chunk. Each chunk is created to not exceed 512 words (approximately 800 tokens). The length of the chunk in number of words was chosen such that running the tokenizer of the pretrained model will result in a sequence that fits within its 1024 token length limit. The header length is chosen to be 128 words (c.f. hyperparameter tuning on the length of header in Appendix~\ref{ssec:addi_results}), representing approximately 25\% of the chunk. The target for every chunk from the same conversation is the complete HPI summary. Contrary to SentBERT, no special care is taken for constructing the summary targets for the chunks (i.e., we use the final desired summary for the conversation) as it is hypothesized the model will learn to only generate information if it is present in the chunk. See Figure~\ref{fig:illus_chunking} for an illustration and Figure~\ref{fig:oak_chunks} for example summaries generated from conversation chunks.

\begin{figure}[tbhp]
    \centering
    \includegraphics[width=0.45\textwidth]{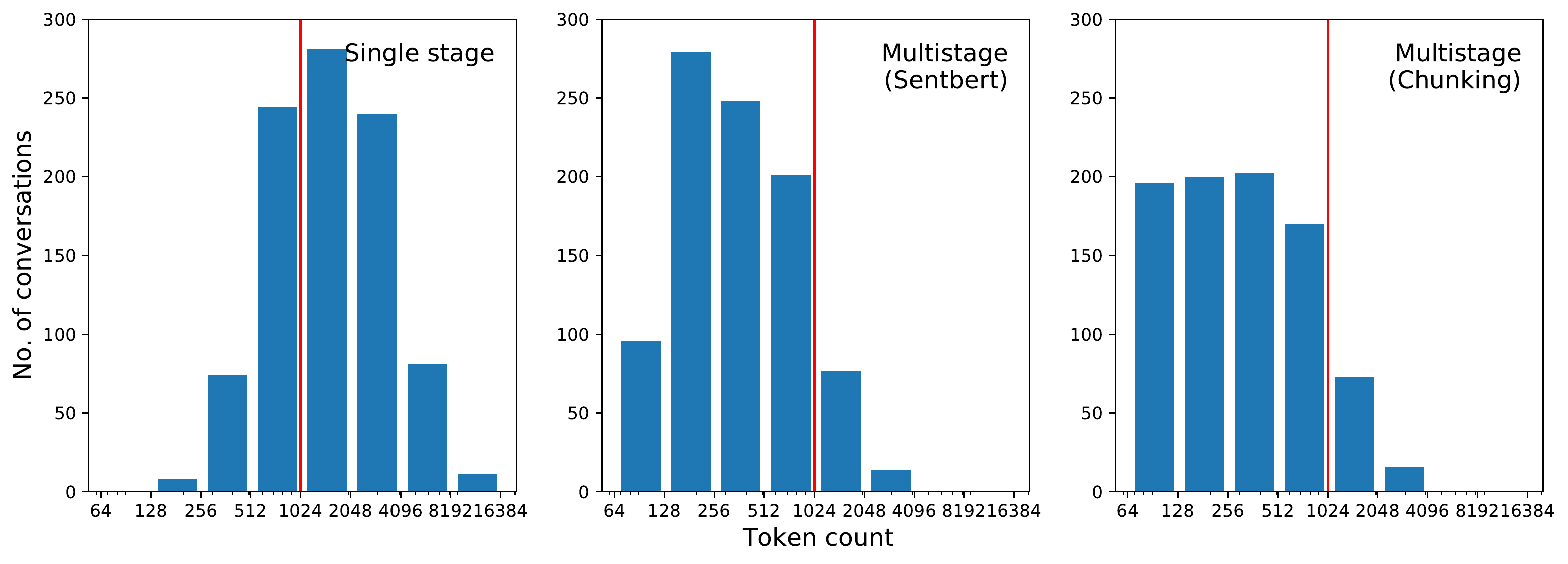}
    \caption{Token count histogram for original conversation (left), 
    input to the second stage of multistage fine-tuning from using SentBERT (middle) and Chunking (right) datasets in the first stage. Vertical lines represent the 1024 token limit after which truncation occurs.}
    \label{fig:tc}
\end{figure}

Our simple multistage approach is proven to be effective in dealing with long conversations. As can be seen in Figure~\ref{fig:tc},  65.3\% of the 939 conversations in the training set exceed the 1024 token limit and 35.5\% exceed 2048 tokens; whereas only less than 10\% of the inputs in the second stage of multistage fine-tuning have to be truncated, regardless of which method we use in the first stage. As we show in Section~\ref{ssec:human_evaluation}, overcoming the truncation problem can help generate summaries that cover information that occurs later in the conversation and have reduced hallucination.

\subsection{Training}
\label{ssec:training}
We leverage the pretrained BART model \cite{lewis2019bart} as our main model for summarization and we choose the model checkpoint pretrained on a BART large model (12 encoder and decoder layers, \textsc{bart-large}, 405 million parameters) as the starting point for all our fine-tuning experiments. For comparison, we also use BigBird \cite{zaheer2020big} with two different model checkpoints: one pretrained using RoBERTa (\textsc{roberta-base}, 155 million parameters) \cite{liu2019roberta} and one from Pegasus (\textsc{pegasus-large}, 575 million parameters) \cite{zhang2020pegasus}. Token limit for all models is set at 1024.\footnote{On Nvidia Titan X Pascal GPU with 12GB memory, we experienced out-of-memory error when using BigBird with a token limit of 2048 or higher, therefore we decided to stay with the default token limit of 1024 and full attention calculation; this means the RoBERTa and Pegasus model checkpoints effectively reduce the BigBird model to \textsc{roberta-base} and \textsc{pegasus-large} models, respectively.} 

The BART experiments are run using fairseq~\cite{ott2019fairseq}, while the BigBird experiments are run using the code released by the authors~\footnote{\url{https://github.com/google-research/bigbird}}. For all fine-tuning experiments, we follow the recommended procedures outlined in their respective repos. We choose the default BPE tokenizer for tokenization with a vocabulary size of 50264. Newline and tab characters in each conversation are replaced by whitespace and no further preprocessing is done. More hyperparameters are shown in Table~\ref{tbl:hyp}. The same hyperparameters are used in both single-stage and multistage fine-tuning.
%
Model checkpoints are saved per epoch. After training, we run model inference on a subset of the development set to pick the checkpoint with the best ROUGE-1 F1 score as the candidate for further evaluation. For single stage fine-tuning on 939 conversations, training is usually finished within 10 epochs.
%


\section{Experiments}
\label{sec:evaluation}


We adopt ROUGE \cite{lin-2004-rouge} as our main evaluation metric. 
Although ROUGE score has limited capability of capturing semantic similarities such as paraphrasing, which is common in abstractive summarization, we still consider it a useful metric for medical summarization due to restricted and highly technical vocabulary used in the medical domain. All references in dev and test set are used in automatic evaluation. 

\begin{table*}[tbhp]
\centering
\begin{tabular}{c|ccc}
\specialrule{0.1em}{0.1em}{0 em}
 & ROUGE-1 F1 & ROUGE-2 F1 & ROUGE-L F1 \\ \hline
BART (large model, single stage) & \textbf{0.3029 (0.4364)} & \textbf{0.1047 (0.1841)} & \textbf{0.3191 (0.4285)} \\
BigBird (\textsc{roberta-base})  & 0.1697 (0.3297) & 0.0633 (0.1662) & 0.1933 (0.3600) \\
BigBird (\textsc{pegasus-large}) & 0.2570 (0.3949) & 0.0822 (0.1889) & 0.2669 (0.3964) \\
\specialrule{0.1em}{0em}{0.1em}
\end{tabular}
\caption{ROUGE evaluation across models on dev set. Numbers in parentheses are "mean-of-best" ROUGE scores. Overall, the results obtained with BigBird were much worse than those obtained with BART, showing the importance of picking an appropriate pretrained model for fine-tuning.}
\label{tbl:rouge_models}
\end{table*}


To address the limitation of ROUGE, we also introduce an automatic concept-based evaluation metric: medically relevant findings are first extracted from both generated and reference summaries by an external NLP system, and then precision, recall, and F1 score are calculated between the two sets of findings. Medical concepts are extracted via one of two systems: our in-house rule-based system and quickUMLS \cite{soldaini2016quickumls}. quickUMLS is a Python implementation of Unified Medical Language System (UMLS)\footnote{\url{https://www.nlm.nih.gov/research/umls/index.html}} that standardizes various health and biomedical vocabularies. It is publicly available, and is capable of extracting a wide scope of medical findings such as symptoms, diseases, medication and procedures. Our rule-based system is a commercial system proven to be effective at capturing symptom-related findings in clinical reports. Example concepts extracted from reference and generated summaries are shown in Appendix~\ref{ssec:running_example}, Figure~\ref{fig:oak_concepts}. False positive error is a major limitation of using those NLP systems, and is more severe with quickUMLS. We therefore implement majority voting to filter medical findings to be included in the reference set: any finding is included only if it is present in at least three human written summaries (or all of them when there are fewer references). The concept-based evaluation based on filtered findings is still susceptible to false positive errors, nevertheless, it provides an alternative to ROUGE as a potentially direct measure of the medical information coverage in generated summaries. Such a measure aligns better with the end-user (i.e., doctors) expectation of the summary quality. We leave research on better metrics for medical summarization as future work. 

For automatic evaluation, we present results on the test set. Results on the development set can be found in Appendix~\ref{ssec:addi_results}. As a more direct approach to quality assessment, we also conduct manual evaluation on a small sample of 10 conversations in the development set.

\subsection{Pretrained model comparison}
\label{ssec:model_comparison}

ROUGE scores for generated summaries across three models: BART, BigBird (RoBERTa), BigBird (Pegasus), are presented in Table~\ref{tbl:rouge_models}. A "mean-of-mean" ROUGE score is calculated by first averaging the scores between the generated summary and all reference summaries for one conversation, and then averaging across conversations. Considering the variance in the length and quality of multiple references, we also calculate a "mean-of-best" ROUGE score: for each conversation, we pick the reference that scores the highest ROUGE-1 F1 with the generated summary and calculate other types of ROUGE scores; we then average the scores across conversations. BART strongly outperformed BigBird with either Roberta or Pegasus checkpoints. Upon manual inspection, we discovered that summaries generated with the BigBird models, or effectively \textsc{roberta-base} and \textsc{pegasus-large}, lack fluency and contain large amounts of repetition with sentences such as \textit{The patient is here for a follow up follow up follow up ...}\footnote{We believe this is not due to different target length settings during inference. We have experimented with 128 and 256 target length for BART as well, and the drop in ROUGE score with shorter target length is no more than $10\%$. Model capacity may not explain the difference in performance either, as BigBird (Pegasus) model contains 40\% more trainable parameters than that of BART.}. We choose to focus on BART in the remaining of the paper.

\subsection{Automatic evaluation}
\label{ssec:multistage_fine-tuning}
ROUGE scores for both single-stage and multistage fine-tuning are shown in Table~\ref{tbl:rouge_bart}. Table~\ref{tbl:concept} shows results for the concept-based evaluation. The Multistage (Chunking) method performs the best by ROUGE metrics, whereas concept-based evaluation leads to mixed results. Differences between the two concept-based evaluations are to be expected considering the different medical findings they cover. It is also worth noting that neither metric moves in unison with ROUGE, we therefore choose to view the three metrics as complementary and providing a more comprehensive interpretation of the quality of the generated summaries. 

Multistage (SentBERT) method does not consistently improve on single-stage training, which could be attributed in part to the mismatch between snippets used in fine-tuning the first stage model and the snippets used for generating single sentence summaries as inputs for the second stage. For example, in Figure~\ref{fig:oak_sentbert} of Appendix~\ref{ssec:addi_results}, we see that some snippets do not contain any noteworthy medical information. The number of such snippets is much larger for the SentBERT method than the Chunking method because of the small span of each snippet and the small stride used to slide over the conversation. This can lead to much noisier inputs for the second stage fine-tuning with more summarizing sentences potentially hallucinating medical contents, that are then unable to be effectively denoised by the second stage model. 



In the last two rows in Table~\ref{tbl:rouge_bart}, we show two baseline evaluations to place other ROUGE scores in context. \textbf{training} computes ROUGE between generated summaries and a set of random target summaries in the training set. The approximate $20\%$ drop in performance provides evidence that the model is not simply memorizing sentences from the training set. 
This is an important concern with medical summarization considering the intrinsic similarity between summaries of the same medical specialty (e.g., similarity among patients with diabetes). \textbf{reference} computes the average ROUGE scores measured among reference summaries. Specifically, for any conversation with multiple references, we do the same ROUGE evaluation used in the rest of the paper by treating each reference in turn as the generated summary and the remaining ones as targets. \textbf{reference} shows the worst scores of all experiments. 
Although this does not guarantee that the generated summaries by the model exceed human performance, we show through the running example and in Section~\ref{ssec:human_evaluation} that model generated summaries can consistently be better than some reference human summaries.

\begin{figure*}[tbhp]
    \centering
    \includegraphics[width=0.32\textwidth]{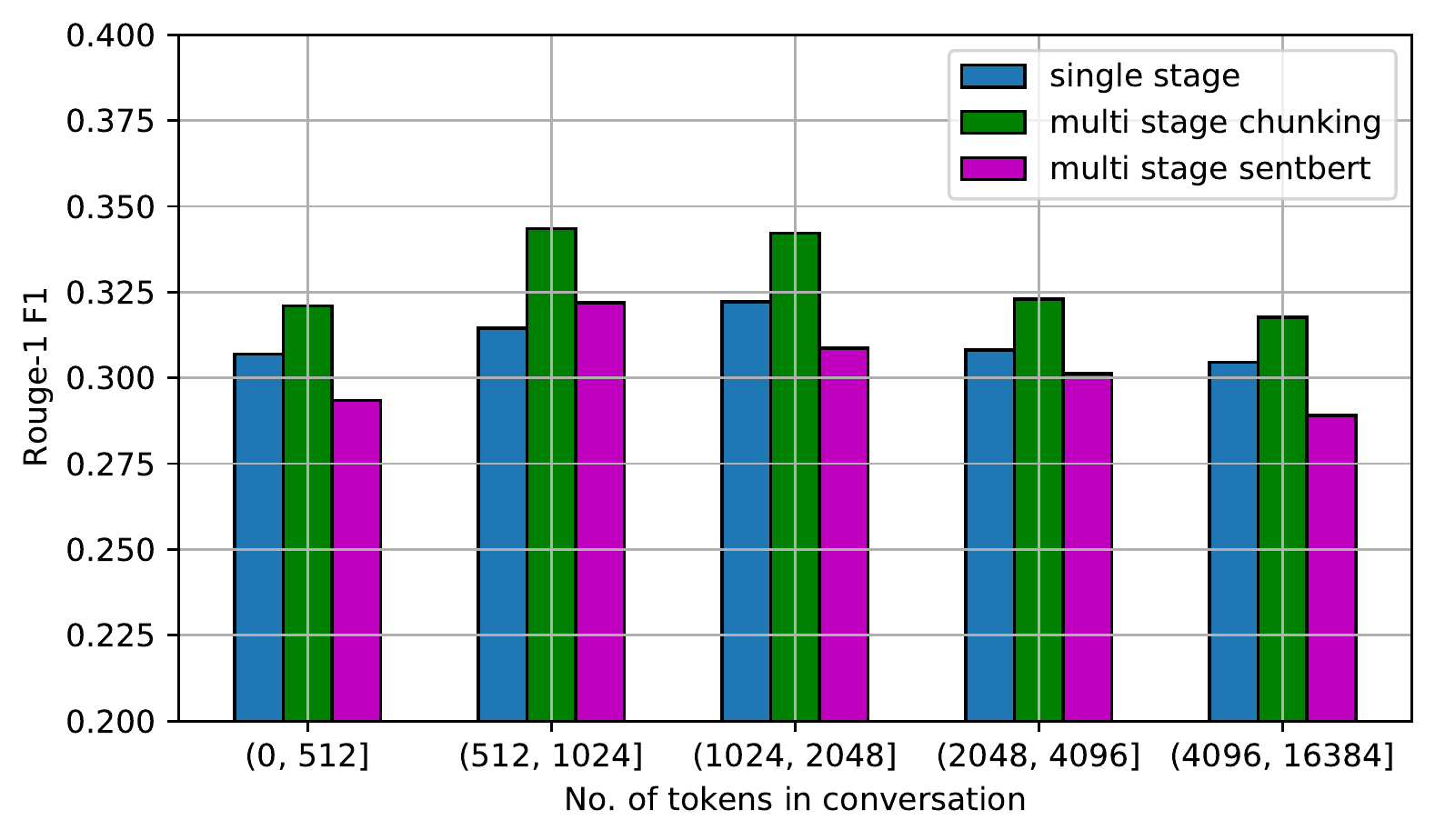} \hspace{0.05cm}
    \includegraphics[width=0.32\textwidth]{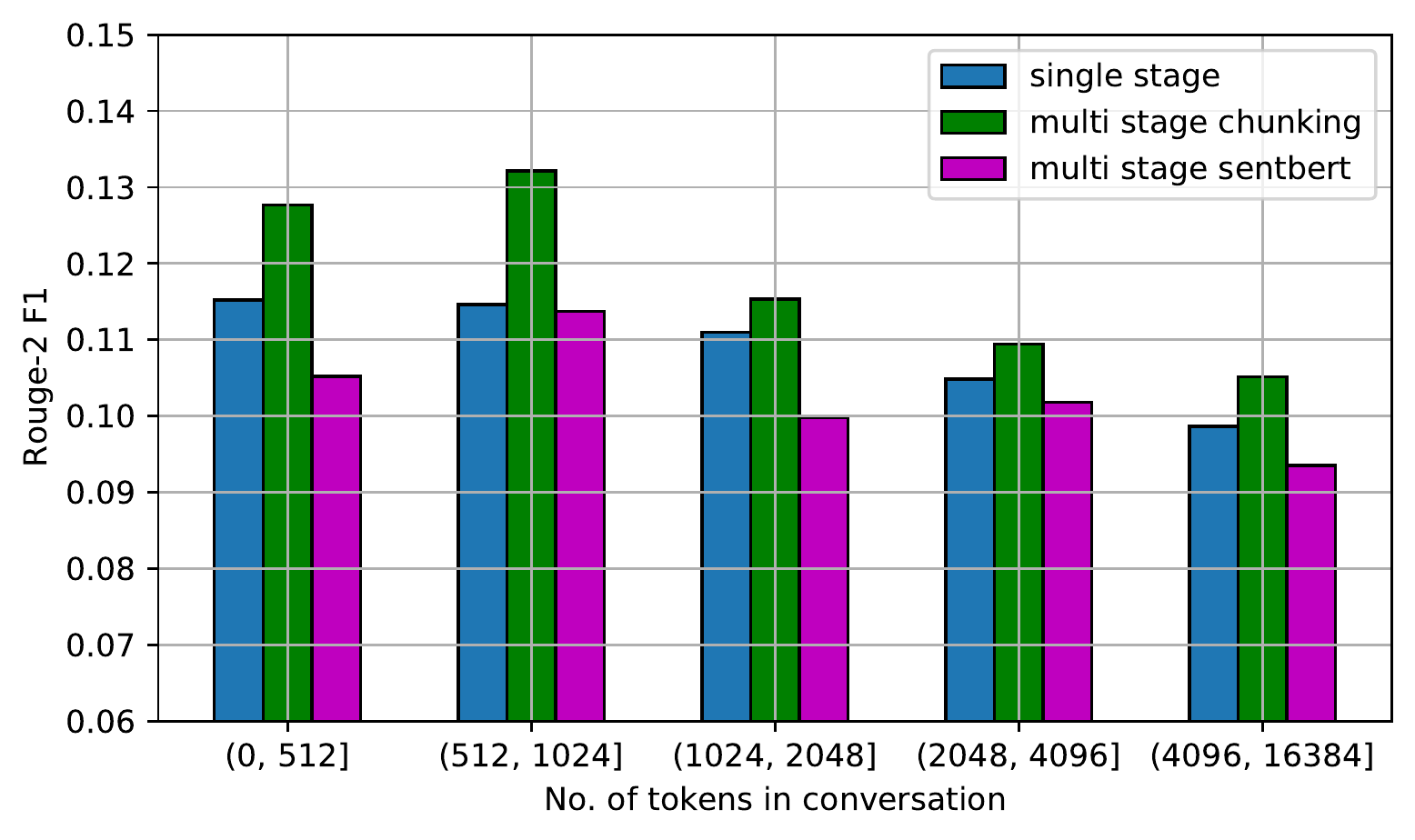} \hspace{0.05cm}
    \includegraphics[width=0.32\textwidth]{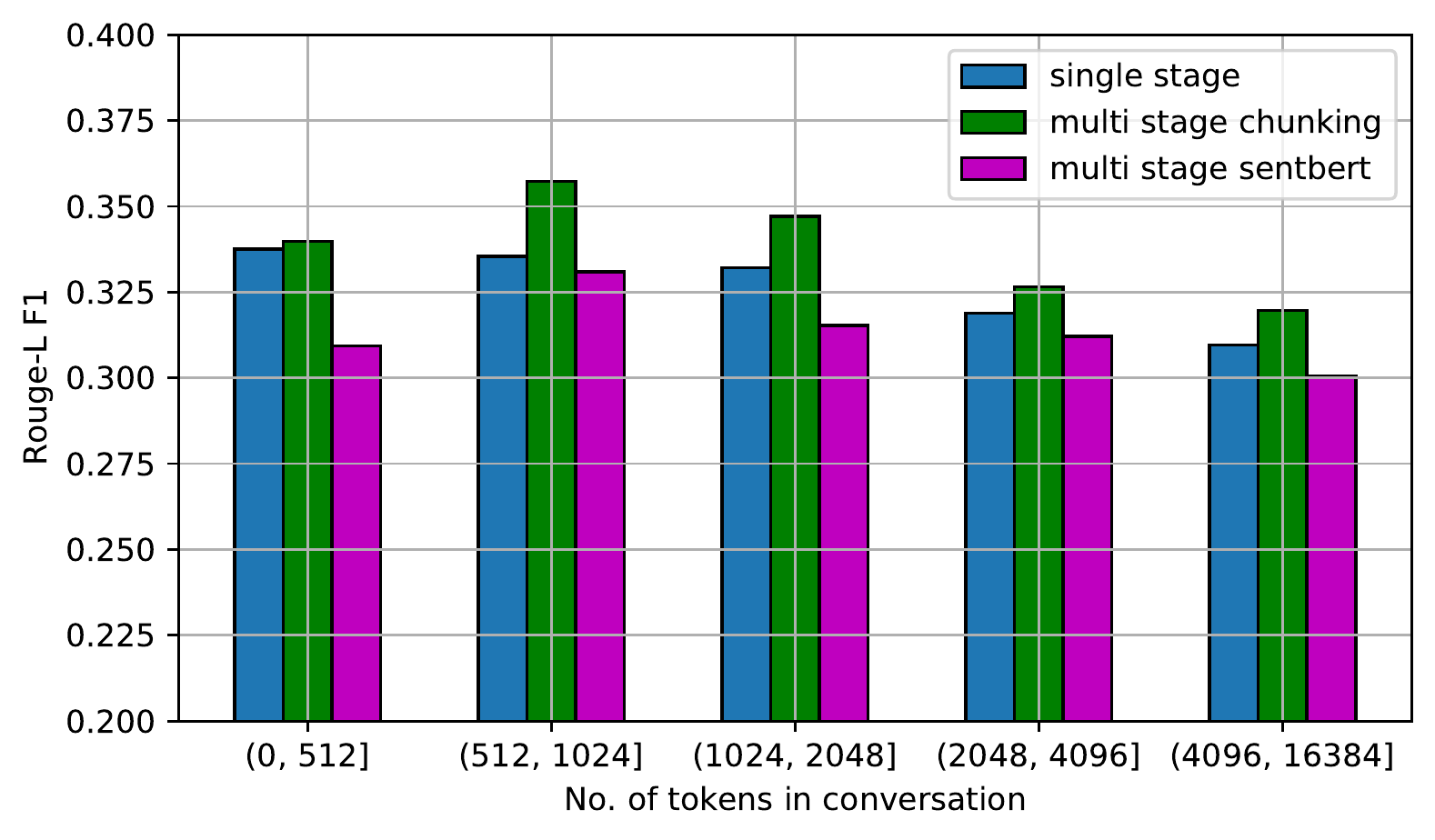}
    
    \includegraphics[width=0.32\textwidth]{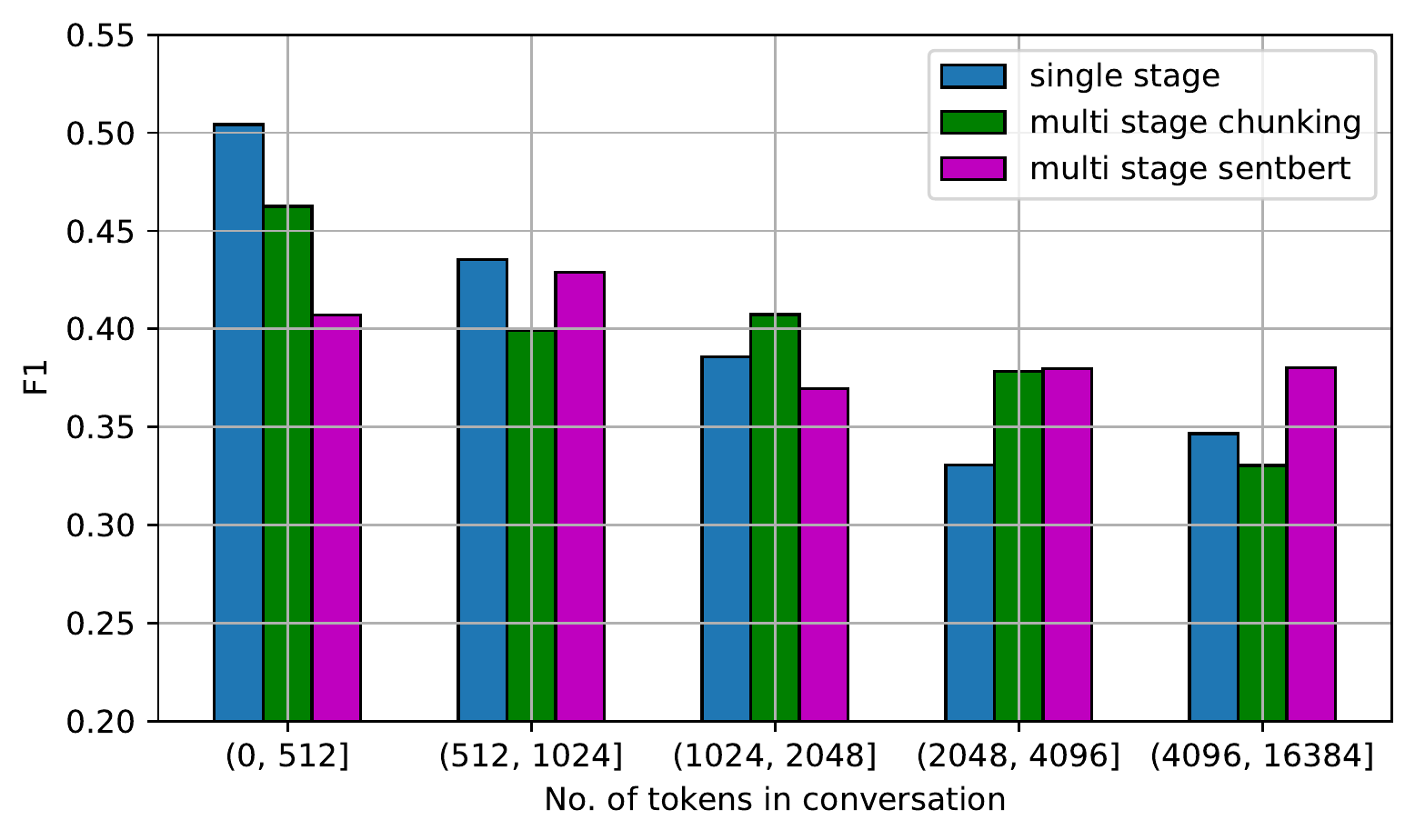} \hspace{0.05cm}
    \includegraphics[width=0.32\textwidth]{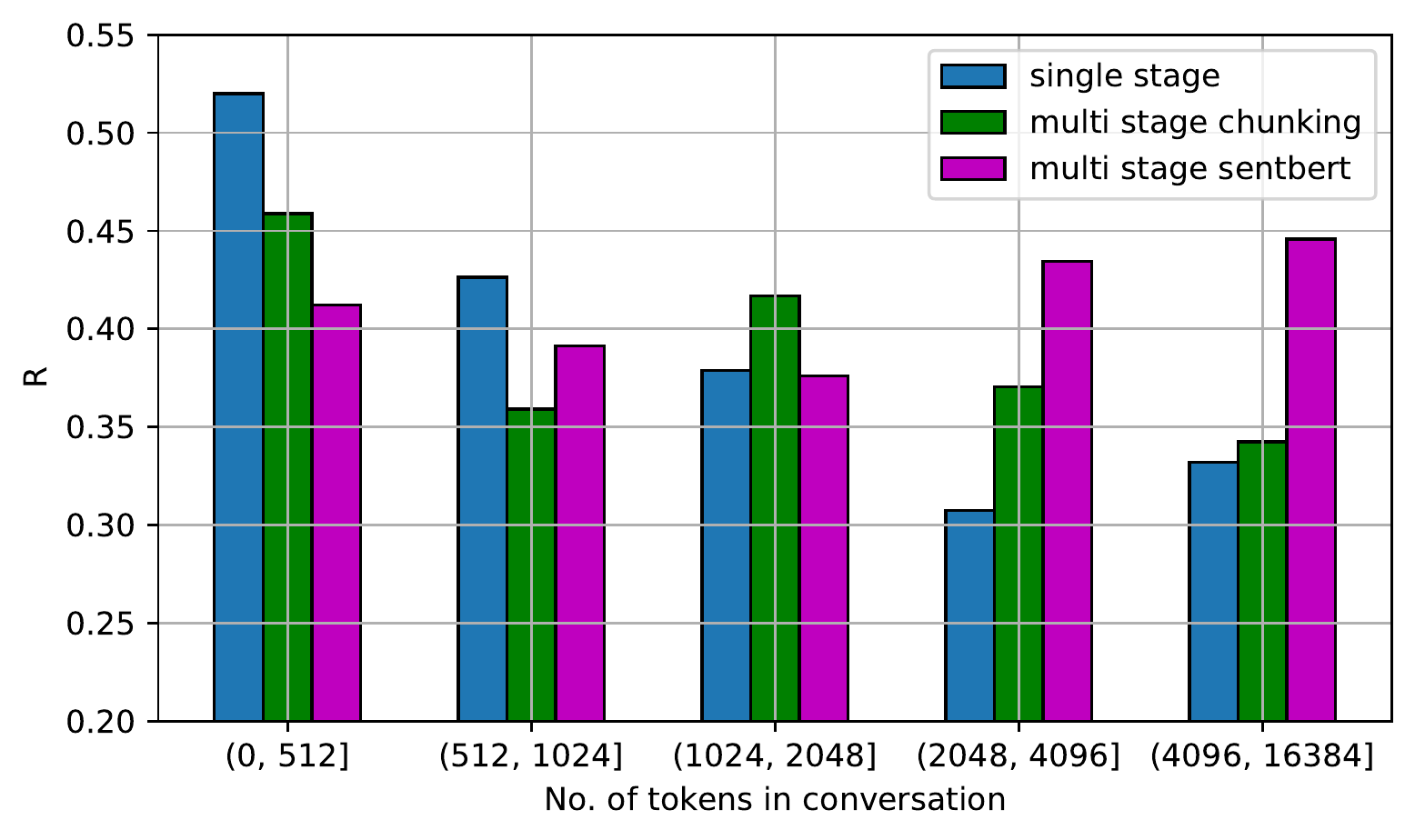} \hspace{0.05cm}
    \includegraphics[width=0.32\textwidth]{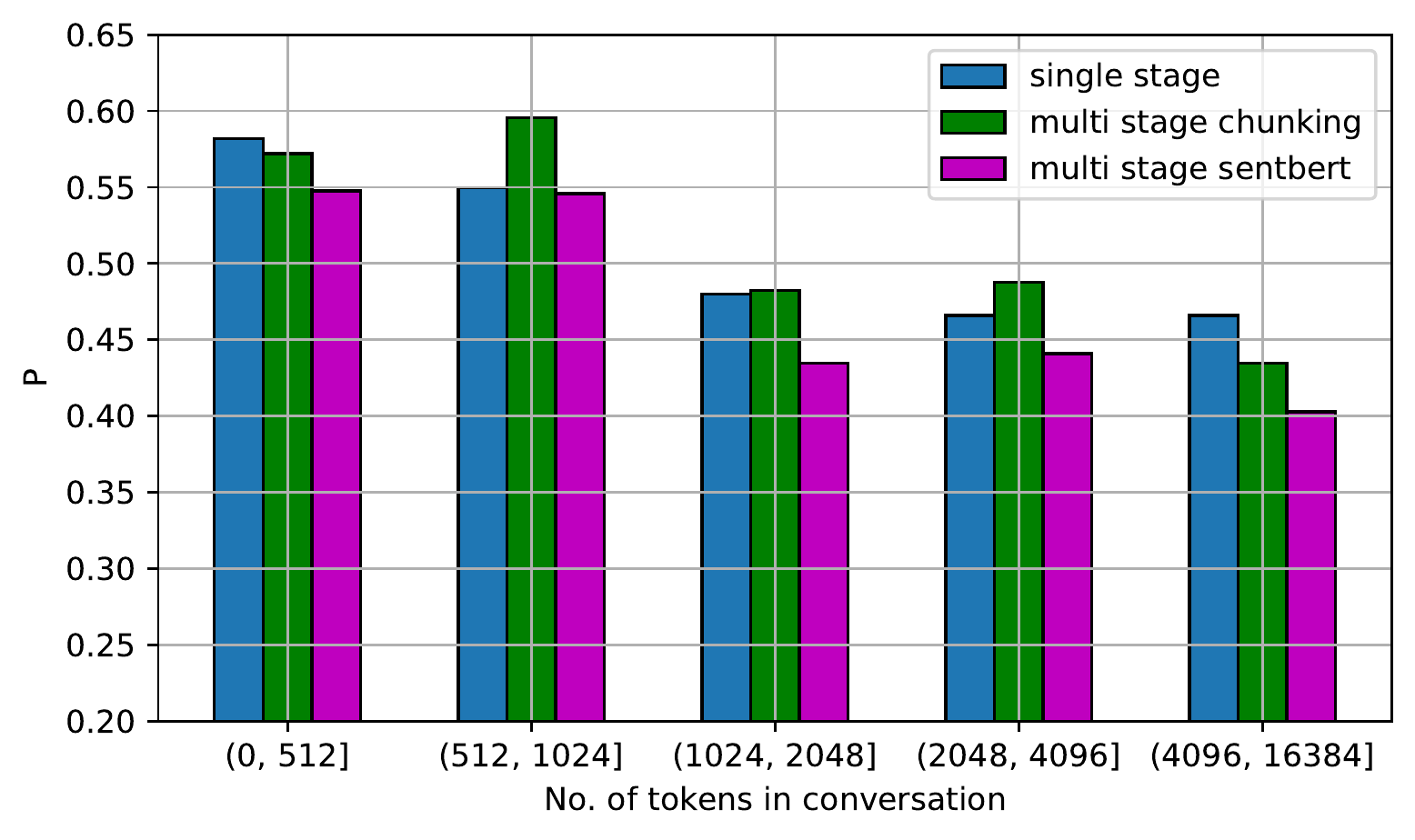}
    \caption{Performance breakdown by number of input tokens. Top row shows "mean-of-mean" ROUGE-1/2/L scores and bottom row displays concept-based F1/R/P using quickUMLS. Results for three models: single stage (\textcolor{blue}{blue}), multistage (SentBERT) (\textcolor{byzantine}{magenta}) and multistage (Chunking) (\textcolor{darkgreen}{green}) are shown. Vertical axis starts at nonzero for better readability.}
    \label{fig:rouge_tokens}
\end{figure*}

Figure~\ref{fig:rouge_tokens} shows the performance breakdown of all three methods by number of input tokens. We group all conversations in the test set into five buckets by their number of input tokens and compare for all methods both the "mean-of-mean" ROUGE scores (top row) and concept-based F1/P/R (bottom row) using quickUMLS. Multistage (Chunking) method outperforms the single stage model consistently across all buckets, even on conversations with fewer than 512 tokens, i.e. conversations that induce only one chunk in the multistage processing; however, the largest improvement in ROUGE score occurs for conversations in the (512, 1024] bucket, which is still within the input token limit of BART model, and we observe similar degradation in ROUGE scores across all three methods as conversation becomes longer. Concept-based evaluation, however, paints a different picture where improvement over single stage method is more significant for conversations beyond the 1024 token limit, which can be largely attributed to improved recall of concepts (see the third and fourth buckets, bottom center plot, Figure \ref{fig:rouge_tokens}). Multistage (SentBERT) method also shows large improvement in concept-based evaluation for very long conversations (larger than 2048 tokens). This suggests both multistage methods lead to more reference medical concepts being generated, which may be favored over minor improvement in ROUGE score in the domain of medical summarization. Multistage (Chunking) method displays the most consistent improvement on conversations in the (1024, 2048] bucket across all types of evaluation metrics, one explanation could be that although multistage training can help circumvent information loss due to truncation, the input to the second stage, namely the concatenation of first stage summaries from all chunks, is also noisy; second stage performance on rewriting such a noisy input could degrade if the level of noise, or the number of first stage summaries, is too large.


\begin{table*}[tbhp]
\centering
\begin{tabular}{c|ccc}
\specialrule{0.1em}{0.1em}{0em}
                      & ROUGE-1 F1               & ROUGE-2 F1               & ROUGE-L F1               \\ \hline
single stage          & 0.3131 (0.4427)          & 0.1097 (0.1819)          & 0.3281 (0.4337)          \\
multistage (Chunking) & \textbf{0.3331 (0.4674)} & \textbf{0.1188 (0.1958)} & \textbf{0.3412 (0.4486)} \\
multistage (SentBERT) & 0.3073 (0.4406)          & 0.1043 (0.1772)          & 0.3170 (0.4218)          \\
training              & 0.2445 (0.3628)          & 0.0588 (0.1198)          & 0.2347 (0.3159)          \\
reference             & 0.2920 (0.4239)          & 0.0852 (0.1638)          & 0.2932 (0.4083)          \\
\specialrule{0.1em}{0em}{0.1em}
\end{tabular}
\caption{ROUGE evaluation for BART fine-tuning on the test set. Values in parentheses are "mean-of-best" scores.}
\label{tbl:rouge_bart}
\end{table*}

\subsection{Human evaluation}
\label{ssec:human_evaluation}

We employ two domain experts to conduct quality evaluation on 10 conversations in the development set. Five short conversations (less than 1024 tokens) and five long conversations (greater than 2048 tokens) are randomly chosen. For each conversation, we include summaries generated from both single-stage and multistage fine-tuning, as well as three reference summaries. One of the three references is selected as the one containing the most symptoms as extracted by our rule-based system (\textbf{reference (max. symp.)} in Table~\ref{tbl:human_score}). The following factors are considered during evaluation: 
\begin{itemize}
    \item \textit{fluency}: How fluent is the text generated?
    \item \textit{relevancy}: Are contents relevant for HPI? 
    \item \textit{missing}: Are any key findings missing?
    \item \textit{hallucination:} Are any findings hallucinated or inaccurate?
    \item \textit{repetition}: Are there repetitive sentences? 
    \item \textit{contradiction}: Are any sentences contradicting each other?
\end{itemize}
Gender mismatch is not considered in the human evaluation as it was observed that, while the model frequently infers the wrong gender pronouns due to the lack of gender information in the transcript, it is sufficient to prefix the conversation with a sentence describing the desired gender for generations to use the correct pronouns. This would allow the development of a system that conditions on self-identified gender information for generation. See Appendix~\ref{ssec:addi_results} for an exploratory experiment.

\begin{table}[tbhp]
\begin{tabular}{c|ccc}
\specialrule{0.1em}{0.1em}{0em}
quickUMLS                                                       & F1              & Precision       & Recall          \\ \hline
single stage                                                    & \textbf{0.4093} & 0.5212          & 0.4009          \\
\begin{tabular}[c]{@{}c@{}}multistage\\ (Chunking)\end{tabular} & 0.4052          & \textbf{0.5316} & 0.3948          \\
\begin{tabular}[c]{@{}c@{}}multistage\\ (SentBERT)\end{tabular} & 0.4001          & 0.4813          & \textbf{0.4166} \\ \specialrule{0.1em}{0em}{0em}
rule-based                                                    & F1              & Precision       & Recall          \\ \hline
single stage                                                    & 0.3617          & \textbf{0.6410} & 0.4112          \\
\begin{tabular}[c]{@{}c@{}}multistage\\ (Chunking)\end{tabular} & \textbf{0.3847} & 0.5951          & 0.4387          \\
\begin{tabular}[c]{@{}c@{}}multistage\\ (SentBERT)\end{tabular} & 0.3673          & 0.5135          & \textbf{0.4622} \\
\specialrule{0.1em}{0em}{0.1em}
\end{tabular}
\caption{Concept-based evaluation on test set.}
\label{tbl:concept}
\end{table}

\paragraph{Inter-rater agreement} We calculate the Pearson's correlation coefficient ($\rho = 0.63$), Kendall rank correlation coefficient ($\tau = 0.51$) and Cohen's kappa ($\kappa = 0.22$) between the two domain experts as measures for inter-rater agreement. The low kappa score should be taken with a grain of salt because of frequent ties in the scores and tie breaking is done somewhat arbitrary during kappa calculation, we therefore focus more on the other two correlation coefficients and consider the agreement between the experts reasonable, but it does reflect the challenge in the consistency of quality evaluation for medical summaries even for experts.

\begin{table*}[tbhp]
\centering
\resizebox{\textwidth}{!}{%
\begin{tabular}{c|cccccc}
\specialrule{0.1em}{0.1em}{0em}
\multicolumn{1}{l|}{}                 & \textit{fluency} & \textit{relevancy} & \textit{missing} & \textit{hallucination} & \textit{repetition} & \textit{contradiction} \\ \hline
single stage                 & 5.0000           & 4.7625             & 3.7375           & 3.8750                 & 5.0000              & 4.6875                 \\
multistage (Chunking)        & 4.9375           & 4.6000             & 3.6875           & 4.2125                 & 4.9250              & 4.7250                 \\
multistage (SentBERT)        & 4.9375           & 4.5375             & 4.0000           & 4.2000                 & 4.8625              & 4.7500                 \\
reference (other)            & 4.8438          & 4.5313            & 2.7813          & 4.9313                & 5.0000                 & 5.0000                    \\
reference (max. symp.) & 4.9375           & 5.0000             & 4.6125           & 4.7250                 & 5.0000              & 5.0000                 \\
\specialrule{0.1em}{0em}{0.1em}
\end{tabular} 
}
\caption{Human evaluation scores on ten conversations. Evaluated on a 5-point scale (higher is better).}
\label{tbl:human_score}
\end{table*}

\paragraph{Qualitative findings} Table~\ref{tbl:human_score} shows the human evaluation scores for all summaries. Scores from the experts are averaged by experts and conversations. \textbf{reference (other)} stands for average score assigned to the other two references in each conversation. The difference in quality across generated and reference summaries are minor in \textit{fluency}, \textit{repetition} and \textit{contradiction}, which indicates the generated summaries are as readable as those written by a human scribe. Generated summaries tend to score lower than the best human reference in \textit{missing} and \textit{hallucination}, with \textit{missing} score being the lowest among all quality factors, suggesting that the fine-tuned models incur more frequently false negative errors. Surprisingly, scores of generated summaries are higher than \textbf{reference (other)} in \textit{relevancy} and \textit{missing} factors. This may be due to the large variability in quality across human references, but does provide encouraging evidence on the potential of using pretrained transformer models towards practical medical dialogue summarization.

Single stage fine-tuning leads to summaries with relevancy comparable to summaries generated by multistage fine-tuning, but with much worse hallucination score. At least among these 10 examples, we do not observe a clear difference in quality between summaries generated by both multistage methods. Hallucination in the single-stage model is more prevalent in longer conversations. For example, in Figure~\ref{fig:oak_summary} in Appendix~\ref{ssec:running_example}, the latter half of the single-stage summary starting from \textit{She has a history of hyperlipidemia...} is largely an hallucination. We believe that this is partly due to the loss of information incurred by truncation (the example conversation contains around 2200 words, or approximately 3500 tokens), resulting in a model that learns to fill in frequently co-occurring information, even if it is not available in the truncated conversation transcript.
Multistage summaries, on the other hand, successfully capture contents beyond the 1024 token limit in the conversation, such as medication like \textit{Cialis}. It is also encouraging to see that the large amount of chitchat  (see, for example, the last chunk in Figure~\ref{fig:oak_chunks}) present in the conversation is largely ignored in the generated summaries from multistage fine-tuning.


\paragraph{Generalization}
As a qualitative comparison with similar work in the field of medical dialogue summarization, we run inference with our fine-tuned models on conversations copied from \citet{krishna2020generating} and \citet{joshi2020dr}. The results are shown in Appendix~\ref{ssec:ood_example}. We include only summaries generated by our single-stage model as all example conversations are well within the 1024 token limit. Summaries generated by the multistage models are of comparable quality. 
The reference summary (Figure~\ref{fig:abridge_ai}) from \citet{krishna2020generating} is a SOAP note \cite{podder2020soap} generated by their best model\footnote{The generated Assessment and Plan (A\&P) section in their paper is not shown because A\&P and HPI sections are largely orthogonal in content.}, which is based on a Pointer-Generator network \cite{pointer-generator}; the gold reference is not provided in the paper. Although generating SOAP notes differs from our summarization task, one can see that our generated summary covers all important medical findings in the reference, with additional findings supported by the conversation (texts highlighted in yellow in Figure~\ref{fig:abridge_ai}). Our generated summary is also much more fluent than the reference paragraph in the "Miscellaneous" section of the reference. One interesting observation is the generation of \textit{hyperlipidemia and diabetes mellitus type 2} in our summary, these findings lack direct evidence from the conversation and may be arguably hallucinations, but it is encouraging that our model successfully infers those diseases from the discussion of insulin and A1c test results in the conversation, which is a very reasonable medical connection that even human scribes are trained to do.
The reference summary for the conversation from \citet{joshi2020dr} (Figure~\ref{fig:dr_summ}) is a gold reference for extractive summarization, with which our abstractive summary shows good agreement. Although some findings in our generated summary, e.g., \textit{Her last two cycles were late by 2 weeks...}, mistakenly mixes concepts mentioned in the conversation, the summary generated by the fine-tuned model has shown promise in generalizing to a medical specialty not present in the training data (OBGYN).

\section{Related work}
\paragraph{Pretrained models} Since the inception of BERT model \cite{devlin2018bert}, the research community has come to the consensus that pretrained, transformer-based models can be effective zero-shot and few-shot learners and there is a constant interest to extend the generalizability and efficiency of such models. \citet{t5} studied the effectiveness of transfer learning of various transformer models and proposed a unified text-to-text framework for all text-based language tasks. \citet{gpt-3} and its earlier versions \cite{gpt-2} showed that it is possible to elicit specific information from the model by providing an appropriate query, or "priming the model". In our work, this is effectively done in annotating each utterance with the corresponding speaker role and breaking the conversation in chunks containing information about the start of the conversation.

\paragraph{Long text summarization} The length of input documents for summarization task is usually limited by the transformer models. One way to break this limit is to overcome the quadratic dependence on the input sequence length of attention calculation, and an abundance of novel transformer architectures with efficient attention modules have been developed in recent years, as explained comprehensively in the survey of \citet{tay2020efficient}. Alternatively, people have been exploring hierarchical structure in the summarization models. \citet{zhang2019hibert} utilized both sentence-level and document-level BERT models to hierarchically encode input documents; \citet{grail2021globalizing} employed BERT model to encode blocks of input text followed by GRU model to integrate encodings across the blocks; \citet{schuller2021windowing} introduced a dynamic windowing approach for a Pointer-Generator network \cite{pointer-generator} to learn to shift between blocks of input as it generates summary sentences sequentially. Our multistage approach for long document summarization introduces a hierarchical structure in the training process (rather than in the model) by going from conversation snippets to a collection of incomplete (pseudo) summaries to a complete summary.

\paragraph{Summarization of medical dialogue} Automatic medical dialogue summarization has started to gain momentum. \citet{krishna2020generating} attempted the generation of complete SOAP note from doctor-patient conversations by first extracting and clustering noteworthy utterances and then leveraging LSTM and transformer models to generate single sentence summary from each cluster. \citet{joshi2020dr} showed that quality of generated summaries can be improved by encouraging copying in pointer-generator network and they also proposed alternative metrics to ROUGE for measuring the medical information coverage. There is also research to address the problems of using ROUGE for evaluating summary quality in the medical domain: \citet{zhang2019optimizing}  explored improving factual correctness of summaries by optimizing ROUGE and concept-based metrics directly as rewards in a reinforcement learning framework of training their summarization model, although a significant difference from our work is that their task was the summarization of radiology reports instead of medical dialogues.

\section{Conclusion}
\label{sec:conclusion}

In this paper, we show the feasibility of summarizing doctor-patient conversation directly from transcripts without an extractive component. We fine-tune various pretrained transformer models for the task of generating the \emph{history of present illness} (HPI) section in a typical medical report from the transcript and achieve surprisingly good performance through pretrained BART models. 
We propose a simple yet general two-stage fine-tuning approach for handling the input length limitation of transformer models: first, a conversation is broken into smaller portions that fit within the length budget of the model and a summarizer is trained on these portions to generate partial summaries; second, we aggregate the generated partial summaries and use them for training a second summarizer to complete the summarization. We show that this approach can help the model pick up medical findings dispersed across long conversations and reduce hallucination compared to single stage fine-tuning.

To the best of our knowledge, our work is the first to show the feasibility of generating fluent summaries directly from doctor-patient conversation transcripts. Of practical concern for medical applications, hallucination and missing information in our generated summaries can be serious problems, nevertheless, we believe our results are encouraging, especially for assisting a scribe in a human-in-the-loop system. We also plan as future work to further explore this task in the aspect of multiple reference summarization and better evaluation metrics that align with quality assessment in the medical domain.


\paragraph{Ethical Considerations} 

Medical conversation summarization inevitably deals with medical data which could potentially contain sensitive information about patients and doctors alike. Careful de-identification for removing all sensitive and identifiable information in the input data is an important tool for privacy protection. We ensured that our data went through a similar process to not reveal any sensitive information (age, name, home address, etc.) about all people involved or mentioned in the conversation. The same de-identified data is also presented to scribes during annotation to ensure no leakage of sensitive information. No information about gender, ethnicity or other discriminating factors are used as a part of our proposed method. 

The intended use of our method is for designing an automatic summarization system aimed at reducing physician and scribe burnout due to the burdersome documentation process required for each medical encounter. 
The most natural application of this technology is not as a replacement for a human scribe, but as an assistant to one. By providing tools that aid a human scribe one can mitigate much of the risk of system failures, such as hallucination. Nonetheless, continued work is required in this area to ensure that both privacy and data accuracy are preserved.

\bibliography{anthology,custom}

\begin{thebibliography}{24}
\expandafter\ifx\csname natexlab\endcsname\relax\def\natexlab#1{#1}\fi

\bibitem[{Amin-Nejad et~al.(2020)Amin-Nejad, Ive, and
  Velupillai}]{amin2020exploring}
Ali Amin-Nejad, Julia Ive, and Sumithra Velupillai. 2020.
\newblock Exploring transformer text generation for medical dataset
  augmentation.
\newblock In \emph{Proceedings of the 12th Language Resources and Evaluation
  Conference}.

\bibitem[{Bird and Klein(2009)}]{nltk}
Edward~Loper Bird, Steven and Ewan Klein. 2009.
\newblock \emph{Natural Language Processing with Python.}
\newblock O'Reilly Media Inc.

\bibitem[{Brown et~al.(2020)Brown, Mann, Ryder, Subbiah, Kaplan, Dhariwal,
  Neelakantan, Shyam, Sastry, Askell et~al.}]{gpt-3}
Tom Brown, Benjamin Mann, Nick Ryder, Melanie Subbiah, Jared Kaplan, Prafulla
  Dhariwal, Arvind Neelakantan, Pranav Shyam, Girish Sastry, Amanda Askell,
  et~al. 2020.
\newblock Language models are few-shot learners.
\newblock \emph{arXiv preprint arXiv:2005.14165}.

\bibitem[{Devlin et~al.(2018)Devlin, Chang, Lee, and
  Toutanova}]{devlin2018bert}
Jacob Devlin, Ming-Wei Chang, Kenton Lee, and Kristina Toutanova. 2018.
\newblock {BERT}: Pre-training of deep bidirectional transformers for language
  understanding.
\newblock \emph{arXiv preprint arXiv:1810.04805}.

\bibitem[{Grail et~al.(2021)Grail, Perez, and Gaussier}]{grail2021globalizing}
Quentin Grail, Julien Perez, and Eric Gaussier. 2021.
\newblock Globalizing {BERT}-based transformer architectures for long document
  summarization.
\newblock In \emph{Proceedings of the 16th Conference of the European Chapter
  of the Association for Computational Linguistics}.

\bibitem[{Huang et~al.(2019)Huang, Altosaar, and
  Ranganath}]{huang2019clinicalbert}
Kexin Huang, Jaan Altosaar, and Rajesh Ranganath. 2019.
\newblock Clinical{BERT}: Modeling clinical notes and predicting hospital
  readmission.
\newblock \emph{arXiv preprint arXiv:1904.05342}.

\bibitem[{Joshi et~al.(2020)Joshi, Katariya, Amatriain, and
  Kannan}]{joshi2020dr}
Anirudh Joshi, Namit Katariya, Xavier Amatriain, and Anitha Kannan. 2020.
\newblock Dr {S}ummarize: Global summarization of medical dialogue by
  exploiting local structures.
\newblock \emph{arXiv preprint arXiv:2009.08666}.

\bibitem[{Krishna et~al.(2020)Krishna, Khosla, Bigham, and
  Lipton}]{krishna2020generating}
Kundan Krishna, Sopan Khosla, Jeffrey Bigham, and Zachary Lipton. 2020.
\newblock Generating {SOAP} notes from doctor-patient conversations.
\newblock \emph{arXiv preprint arXiv:2005.01795}.

\bibitem[{Lewis et~al.(2019)Lewis, Liu, Goyal, Ghazvininejad, Mohamed, Levy,
  Stoyanov, and Zettlemoyer}]{lewis2019bart}
Mike Lewis, Yinhan Liu, Naman Goyal, Marjan Ghazvininejad, Abdelrahman Mohamed,
  Omer Levy, Ves Stoyanov, and Luke Zettlemoyer. 2019.
\newblock {BART}: Denoising sequence-to-sequence pre-training for natural
  language generation, translation, and comprehension.
\newblock \emph{arXiv preprint arXiv:1910.13461}.

\bibitem[{Lin(2004)}]{lin-2004-rouge}
Chin-Yew Lin. 2004.
\newblock {ROUGE}: A package for automatic evaluation of summaries.
\newblock In \emph{Text Summarization Branches Out}.

\bibitem[{Liu et~al.(2019)Liu, Ott, Goyal, Du, Joshi, Chen, Levy, Lewis,
  Zettlemoyer, and Stoyanov}]{liu2019roberta}
Yinhan Liu, Myle Ott, Naman Goyal, Jingfei Du, Mandar Joshi, Danqi Chen, Omer
  Levy, Mike Lewis, Luke Zettlemoyer, and Veselin Stoyanov. 2019.
\newblock Roberta: A robustly optimized bert pretraining approach.
\newblock \emph{arXiv preprint arXiv:1907.11692}.

\bibitem[{Ott et~al.(2019)Ott, Edunov, Baevski, Fan, Gross, Ng, Grangier, and
  Auli}]{ott2019fairseq}
Myle Ott, Sergey Edunov, Alexei Baevski, Angela Fan, Sam Gross, Nathan Ng,
  David Grangier, and Michael Auli. 2019.
\newblock fairseq: A fast, extensible toolkit for sequence modeling.
\newblock \emph{arXiv preprint arXiv:1904.01038}.

\bibitem[{Podder et~al.(2020)Podder, Lew, and Ghassemzadeh}]{podder2020soap}
Vivek Podder, Valerie Lew, and Sassan Ghassemzadeh. 2020.
\newblock Soap notes.
\newblock \emph{StatPearls [Internet]}.

\bibitem[{Radford et~al.(2019)Radford, Wu, Child, Luan, Amodei, and
  Sutskever}]{gpt-2}
Alec Radford, Jeffrey Wu, Rewon Child, David Luan, Dario Amodei, and Ilya
  Sutskever. 2019.
\newblock Language models are unsupervised multitask learners.
\newblock \emph{OpenAI blog}.

\bibitem[{Raffel et~al.(2019)Raffel, Shazeer, Roberts, Lee, Narang, Matena,
  Zhou, Li, and Liu}]{t5}
Colin Raffel, Noam Shazeer, Adam Roberts, Katherine Lee, Sharan Narang, Michael
  Matena, Yanqi Zhou, Wei Li, and Peter Liu. 2019.
\newblock Exploring the limits of transfer learning with a unified text-to-text
  transformer.
\newblock \emph{arXiv preprint arXiv:1910.10683}.

\bibitem[{Reimers and Gurevych(2019)}]{reimers2019sentence}
Nils Reimers and Iryna Gurevych. 2019.
\newblock Sentence-{BERT}: Sentence embeddings using siamese {BERT}-networks.
\newblock \emph{arXiv preprint arXiv:1908.10084}.

\bibitem[{Sch{\"u}ller et~al.(2021)Sch{\"u}ller, Wilhelm, Kreiling, and
  Glava{\v{s}}}]{schuller2021windowing}
Leon Sch{\"u}ller, Florian Wilhelm, Nico Kreiling, and Goran Glava{\v{s}}.
  2021.
\newblock Windowing models for abstractive summarization of long texts.
\newblock In \emph{European Conference on Information Retrieval}, pages
  384--392. Springer.

\bibitem[{See et~al.(2017)See, Liu, and Manning}]{pointer-generator}
Abigail See, Peter~J Liu, and Christopher~D Manning. 2017.
\newblock Get to the point: Summarization with pointer-generator networks.
\newblock \emph{arXiv preprint arXiv:1704.04368}.

\bibitem[{Soldaini and Goharian(2016)}]{soldaini2016quickumls}
Luca Soldaini and Nazli Goharian. 2016.
\newblock Quickumls: a fast, unsupervised approach for medical concept
  extraction.
\newblock In \emph{MedIR workshop, sigir}, pages 1--4.

\bibitem[{Tay et~al.(2020)Tay, Dehghani, Bahri, and Metzler}]{tay2020efficient}
Yi~Tay, Mostafa Dehghani, Dara Bahri, and Donald Metzler. 2020.
\newblock Efficient transformers: A survey.
\newblock \emph{arXiv preprint arXiv:2009.06732}.

\bibitem[{Zaheer et~al.(2020)Zaheer, Guruganesh, Dubey, Ainslie, Alberti,
  Ontanon, Pham, Ravula, Wang, Yang et~al.}]{zaheer2020big}
Manzil Zaheer, Guru Guruganesh, Avinava Dubey, Joshua Ainslie, Chris Alberti,
  Santiago Ontanon, Philip Pham, Anirudh Ravula, Qifan Wang, Li~Yang, et~al.
  2020.
\newblock Big bird: Transformers for longer sequences.
\newblock \emph{arXiv preprint arXiv:2007.14062}.

\bibitem[{Zhang et~al.(2020)Zhang, Zhao, Saleh, and Liu}]{zhang2020pegasus}
Jingqing Zhang, Yao Zhao, Mohammad Saleh, and Peter Liu. 2020.
\newblock Pegasus: Pre-training with extracted gap-sentences for abstractive
  summarization.
\newblock In \emph{International Conference on Machine Learning}, pages
  11328--11339. PMLR.

\bibitem[{Zhang et~al.(2019{\natexlab{a}})Zhang, Wei, and
  Zhou}]{zhang2019hibert}
Xingxing Zhang, Furu Wei, and Ming Zhou. 2019{\natexlab{a}}.
\newblock {HIBERT}: Document level pre-training of hierarchical bidirectional
  transformers for document summarization.
\newblock \emph{arXiv preprint arXiv:1905.06566}.

\bibitem[{Zhang et~al.(2019{\natexlab{b}})Zhang, Merck, Tsai, Manning, and
  Langlotz}]{zhang2019optimizing}
Yuhao Zhang, Derek Merck, Emily~Bao Tsai, Christopher~D Manning, and Curtis~P
  Langlotz. 2019{\natexlab{b}}.
\newblock Optimizing the factual correctness of a summary: A study of
  summarizing radiology reports.
\newblock \emph{arXiv preprint arXiv:1911.02541}.

\end{thebibliography}
\bibliographystyle{acl_natbib}

\clearpage
\appendix
\counterwithin{figure}{section}
\counterwithin{table}{section}

\section{Appendix}
\label{sec:appendix}

\subsection{Dataset Statistics}
\label{ssec:data_stats}
\begin{figure}[tbhp]
    \centering
    \includegraphics[width=0.45\textwidth]{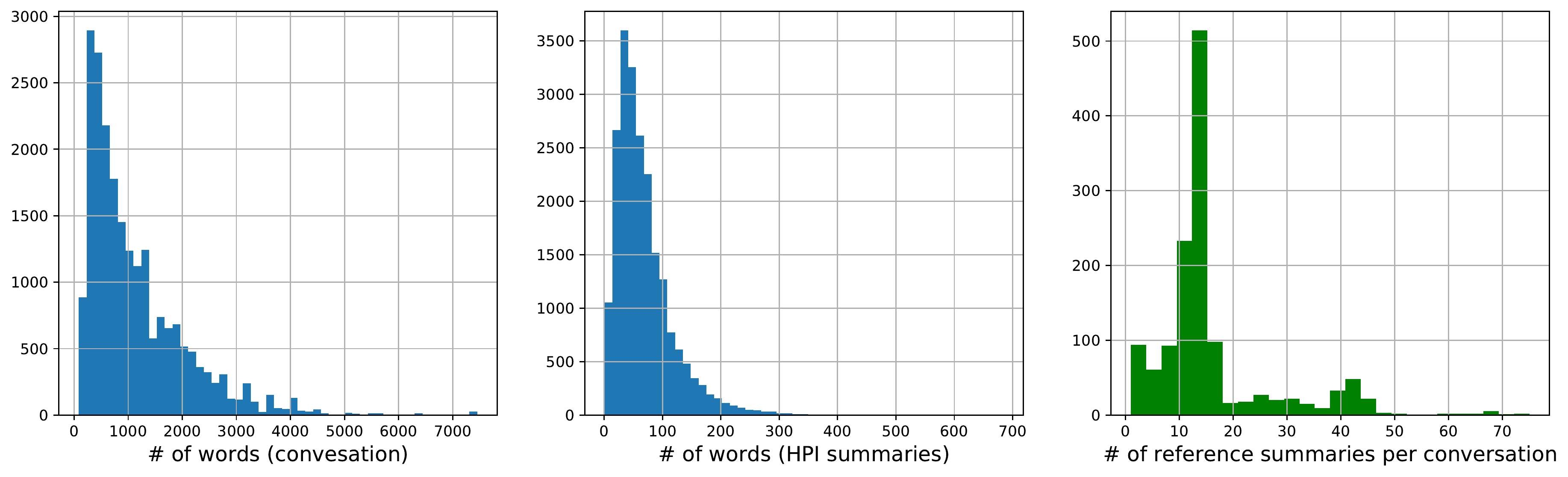}
    \caption{Dataset statistics: word count in conversation (left); word count in HPI summaries (middle); no. of reference summaries (right).}
    \label{fig:wc_conv}
\end{figure}

See Figure~\ref{fig:wc_conv} for statistics on word count and number of reference summaries in the dataset.

\subsection{Hyperparameters}
\label{ssec:hyp}

Table~\ref{tbl:hyp} lists the typical hyperparameters used in training and inference for both BART and BigBird models. BART models are trained on AWS Sagemaker instances with a single Nvidia V100 GPU (16 GB Memory); BigBird models are fine-tuned on our internal server with a single NVidia Titan X Pascal GPU (12GB memory).

\begin{table*}[tbhp]
\begin{tabular}{l|c|c|c}
\specialrule{0.1em}{0.1em}{0em}
\textbf{Parameter} & \textbf{BART} & \textbf{BigBird (RoBERTa)}  & \textbf{BigBird (Pegasus)}  \\ \hline
Learning rate & $2.5 \times 10^{-5}$ & $1 \times 10^{-5}$ & $1 \times 10^{-4}$  \\ \hline
LR schedule & \begin{tabular}[c]{@{}l@{}}polynomial \\ 200 steps warmup\\ 30000 steps total\end{tabular} & \begin{tabular}[c]{@{}l@{}}Square root decay \\ 100 steps linear warmup\\ 30000 steps total\end{tabular}& 
\begin{tabular}[c]{@{}l@{}}Square root decay \\ 100 steps linear warmup\\ 30000 steps total\end{tabular}\\ \hline
Batch size   & 1 ($\times 8$) & 1 & 1  \\ \hline
Optimizer              & Adam       & Adam              &  Adafactor         \\ \hline
Dropout                & 0.1                            & 0.1               & 0.1               \\ \hline
\begin{tabular}[c]{@{}l@{}}Early stopping \\ monitor \end{tabular}        & dev set NLL loss               & dev set NLL loss  & dev set NLL loss               \\ \hline
\begin{tabular}[c]{@{}l@{}}Early stopping \\ patience \end{tabular}        & 3               & 3  & 3               \\ \hline
\begin{tabular}[c]{@{}l@{}}Beam search \\ \# of hypotheses\end{tabular} & 4 & 5 & 5 \\ \hline
\begin{tabular}[c]{@{}l@{}}Beam search \\ maximum generation length \\ (\# of tokens)\end{tabular} & 512 & 256 & 256 \\ \hline
\begin{tabular}[c]{@{}l@{}}Beam search \\ length penalty\end{tabular} & 0.2 & 0.7 & 0.7 \\ \hline
\specialrule{0.1em}{0em}{0.1em}
\end{tabular}
\caption{Hyperparameter settings. $\times 8$ in batch size setting specifies no. of updates used in gradient accumulation.}
\label{tbl:hyp}
\end{table*}

\subsection{Running Example}
\label{ssec:running_example}
\begin{figure*}[tbhp]
    \centering
    \includegraphics[width=1.0\textwidth]{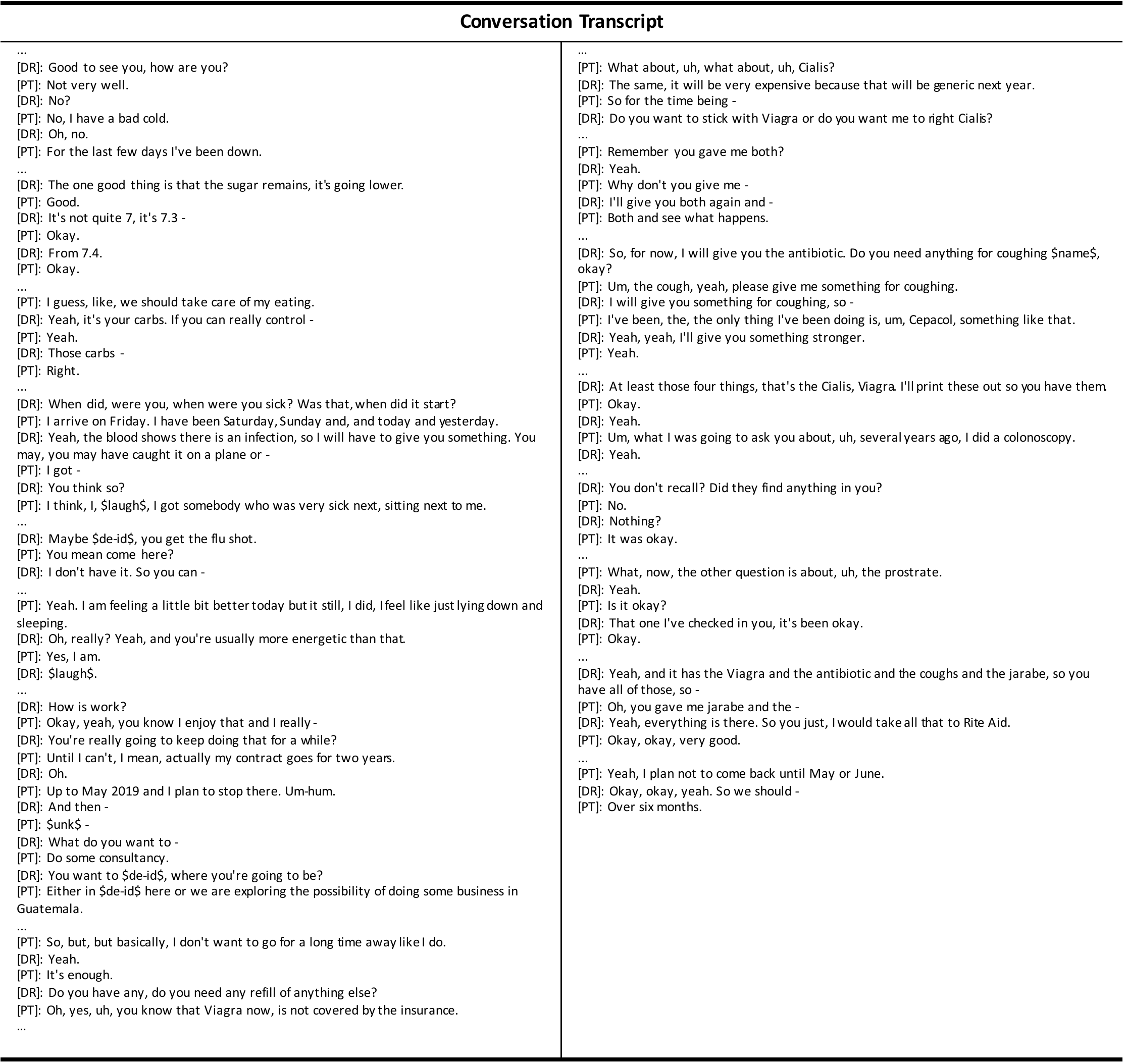}
    \caption{Excerpt from example conversation in the dataset. Due to privacy requirement, only portions of conversation relevant to the summaries are shown. Snippets in the left panel are within the 1024 token limit and those in the right panel are beyond the limit.}
    \label{fig:oak_example}
\end{figure*}

\begin{figure*}[tbhp]
    \centering
    \includegraphics[width=1.0\textwidth]{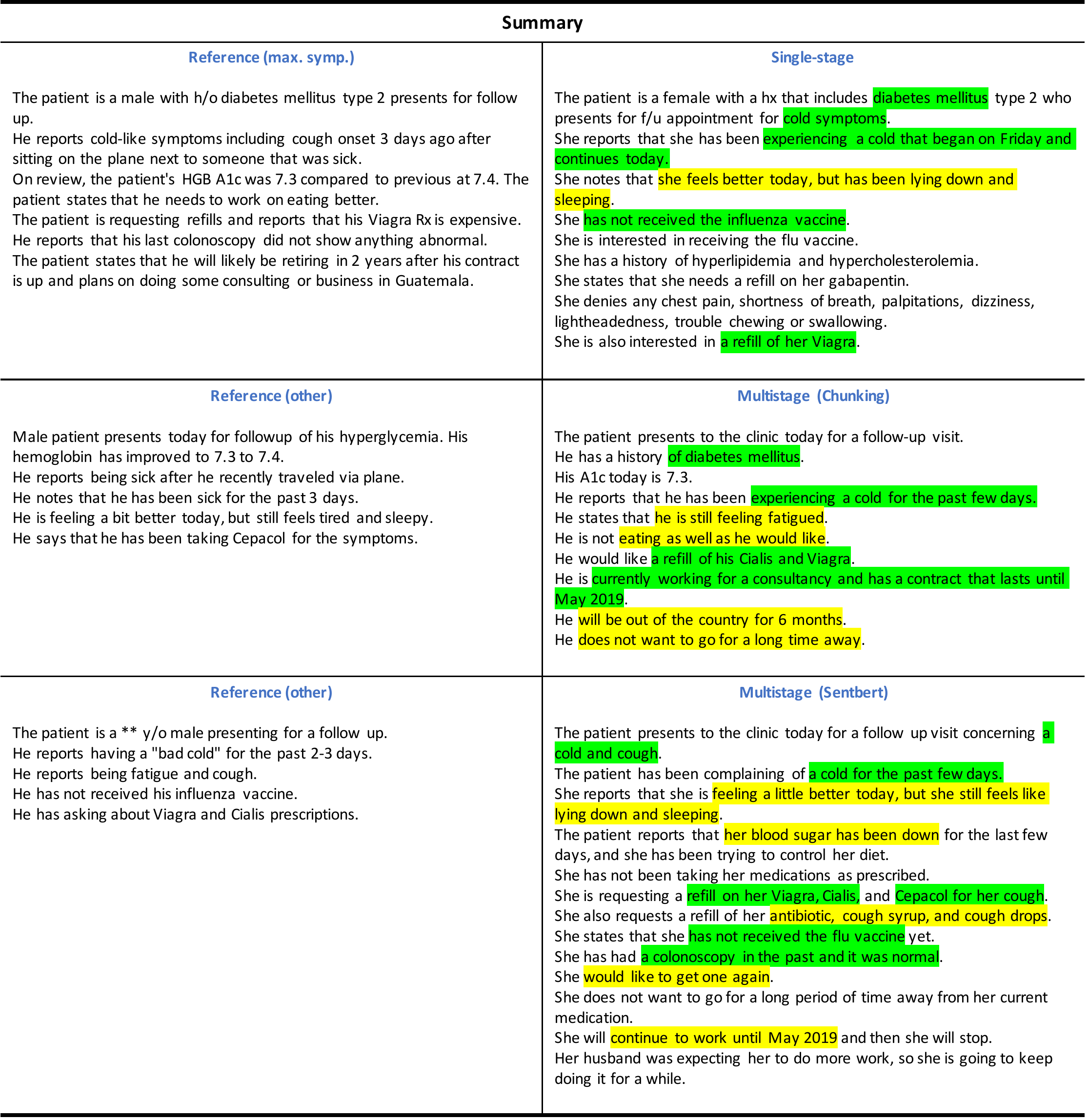}
    \caption{BART generated summaries and references. Text in \textcolor{green}{green} highlights medical findings present in at least one reference summary; text with \textcolor{yellow}{yellow} highlighting shows findings not in reference but are supported by the conversation.}
    \label{fig:oak_summary}
\end{figure*}

\begin{figure*}[tbhp]
    \centering
    \includegraphics[width=0.95\textwidth]{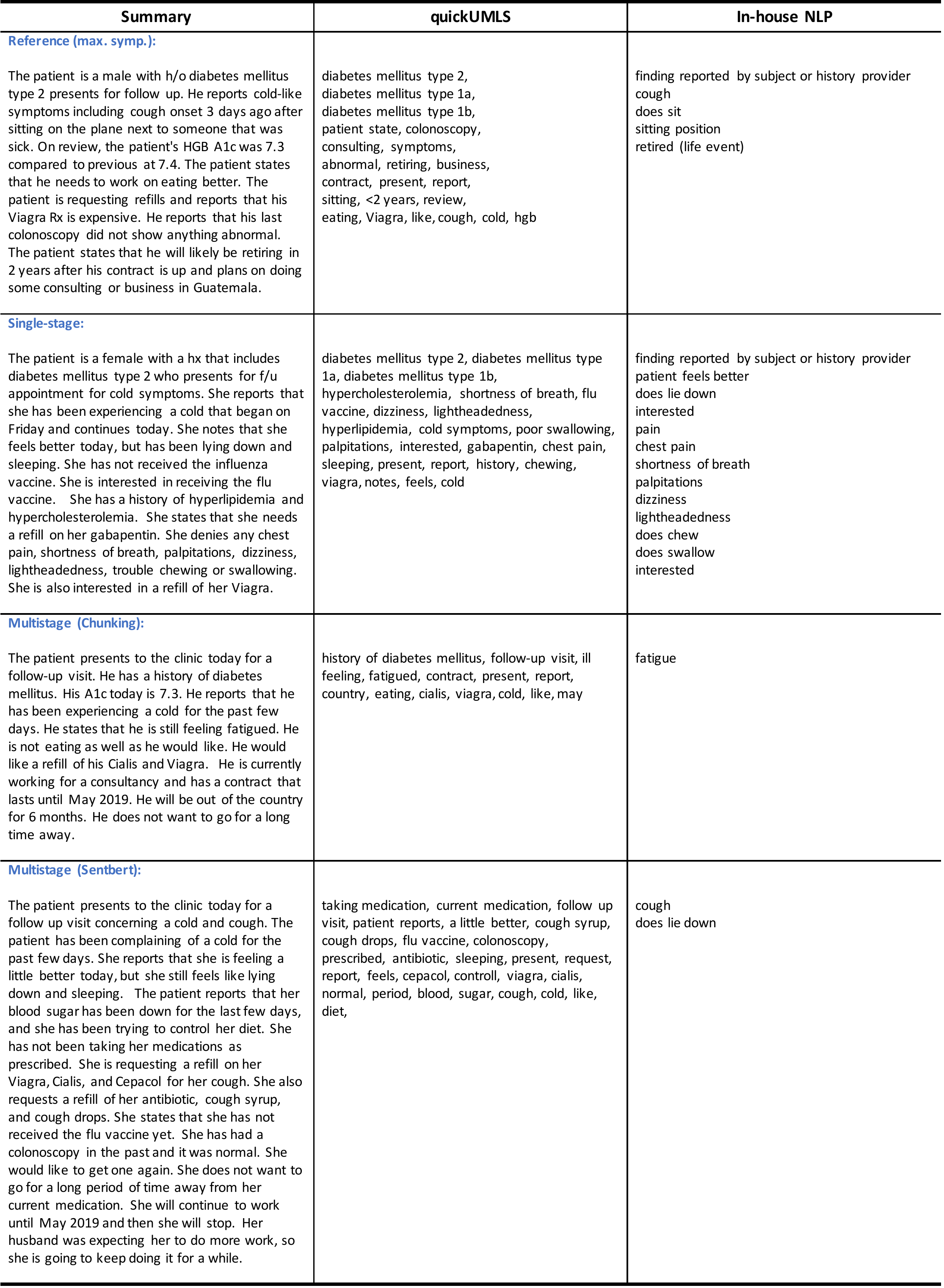}
    \caption{Medical concepts extracted from summaries by quickUMLS and our rule-based system. UMLS findings (second column) are separated by commas, and rule-based findings (third column) are shown on separate lines. In order to control the generation of false positive concepts, we choose to consider for evaluation only clinical findings (symptoms) extracted by the In-house NLP system; disorders (e.g. diabetes mellitus), medications and clinical procedures (e.g., colonoscopy) are ignored, which are concepts of lower priority in the HPI section of an EHR report.}
    \label{fig:oak_concepts}
\end{figure*}

We showcase an example conversation, the corresponding reference and model generated summaries, and extracted medical findings by quickUMLS and our rule-based system in Figure~\ref{fig:oak_example}-\ref{fig:oak_concepts}. These examples are referred to throughout the paper.

\begin{figure*}[tbhp]
    \centering
    \includegraphics[width=1.0\textwidth]{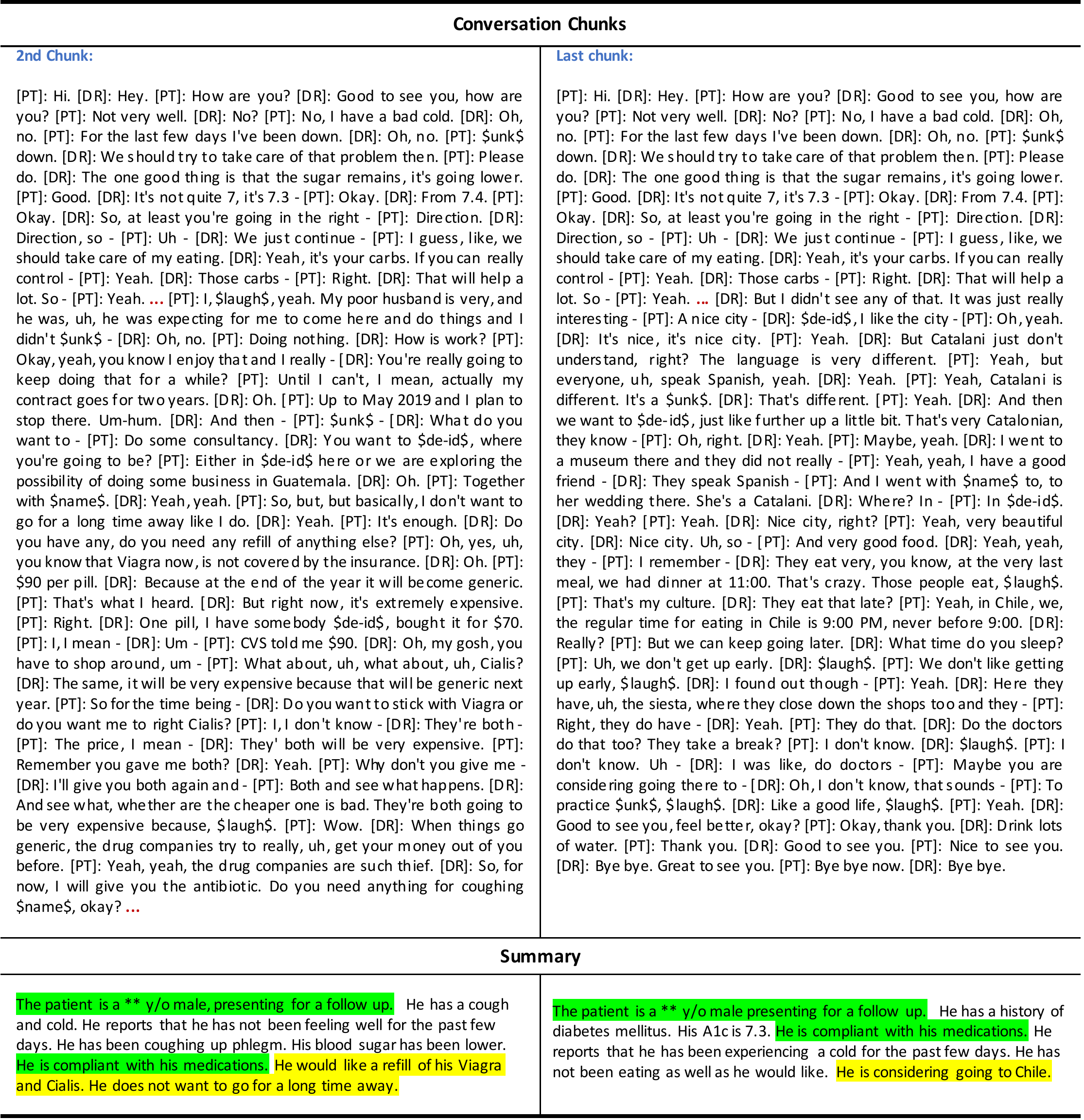}
    \caption{Example conversation chunks and generated summary in the first stage of multistage fine-tuning with Chunking method. Text highlighted in \textcolor{green}{green} are sentences common to both summaries. Text in  \textcolor{yellow}{yellow} marks sentences that are supported by the "body" part of each chunk. Note that \textcolor{red}{...} is used in each chunk to mask the rest of the conversation.}
    \label{fig:oak_chunks}
\end{figure*}

\begin{figure*}[tbhp]
    \centering
    \includegraphics[width=1.0\textwidth]{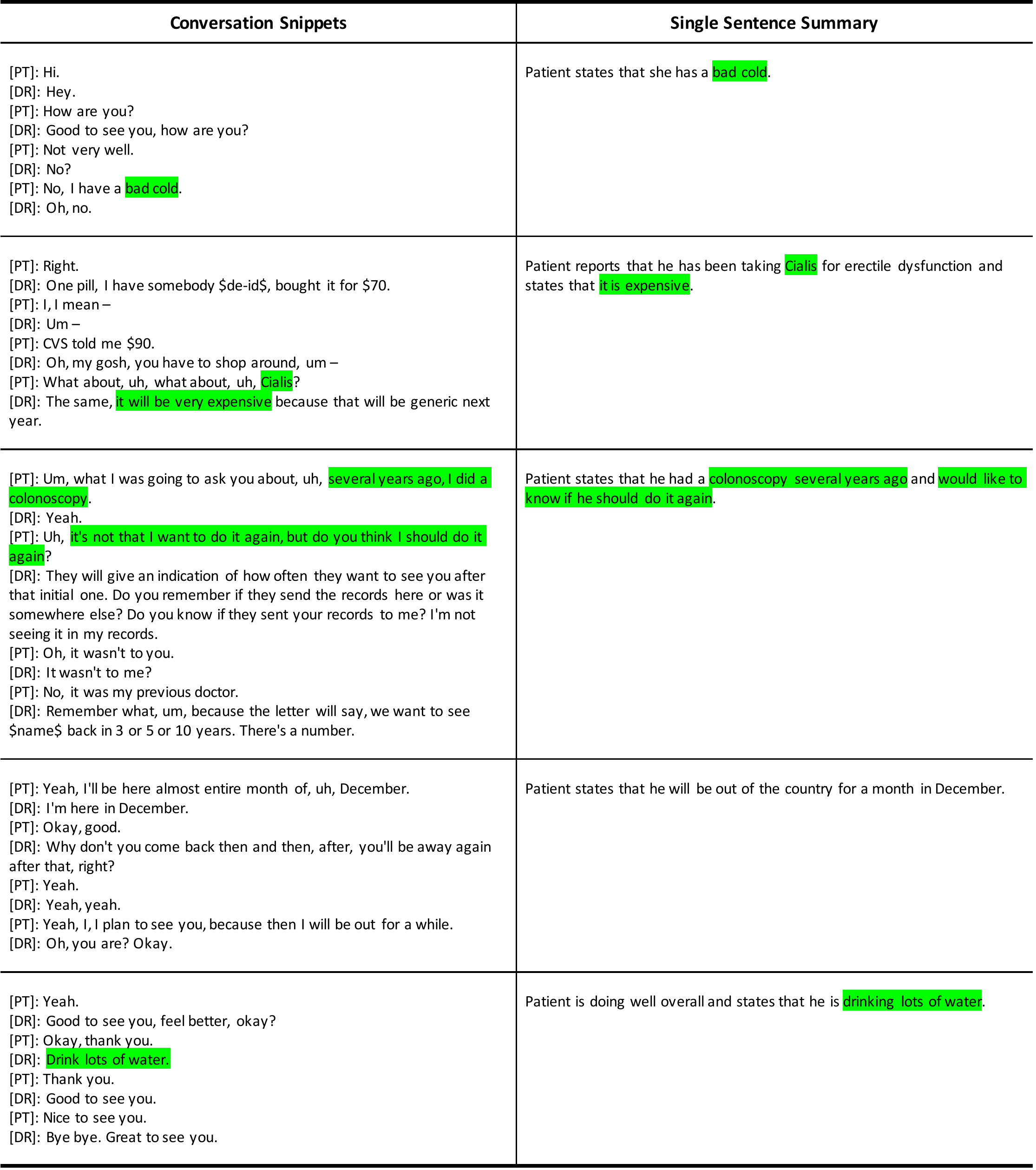}
    \caption{Example conversation snippets and single sentence summary used in the first stage of multistage fine-tuning with the SentBERT method. Snippets chosen approximately equi-distance from each other from the beginning to the end of the conversation. Text highlighted in \textcolor{green}{green} show generated contents and the supported text in the corresponding snippets.}
    \label{fig:oak_sentbert}
\end{figure*}

\subsection{Inference on Out-of-dataset Examples}
\label{ssec:ood_example}
\begin{figure*}[tbhp]
    \centering
    \includegraphics[width=1.0\textwidth]{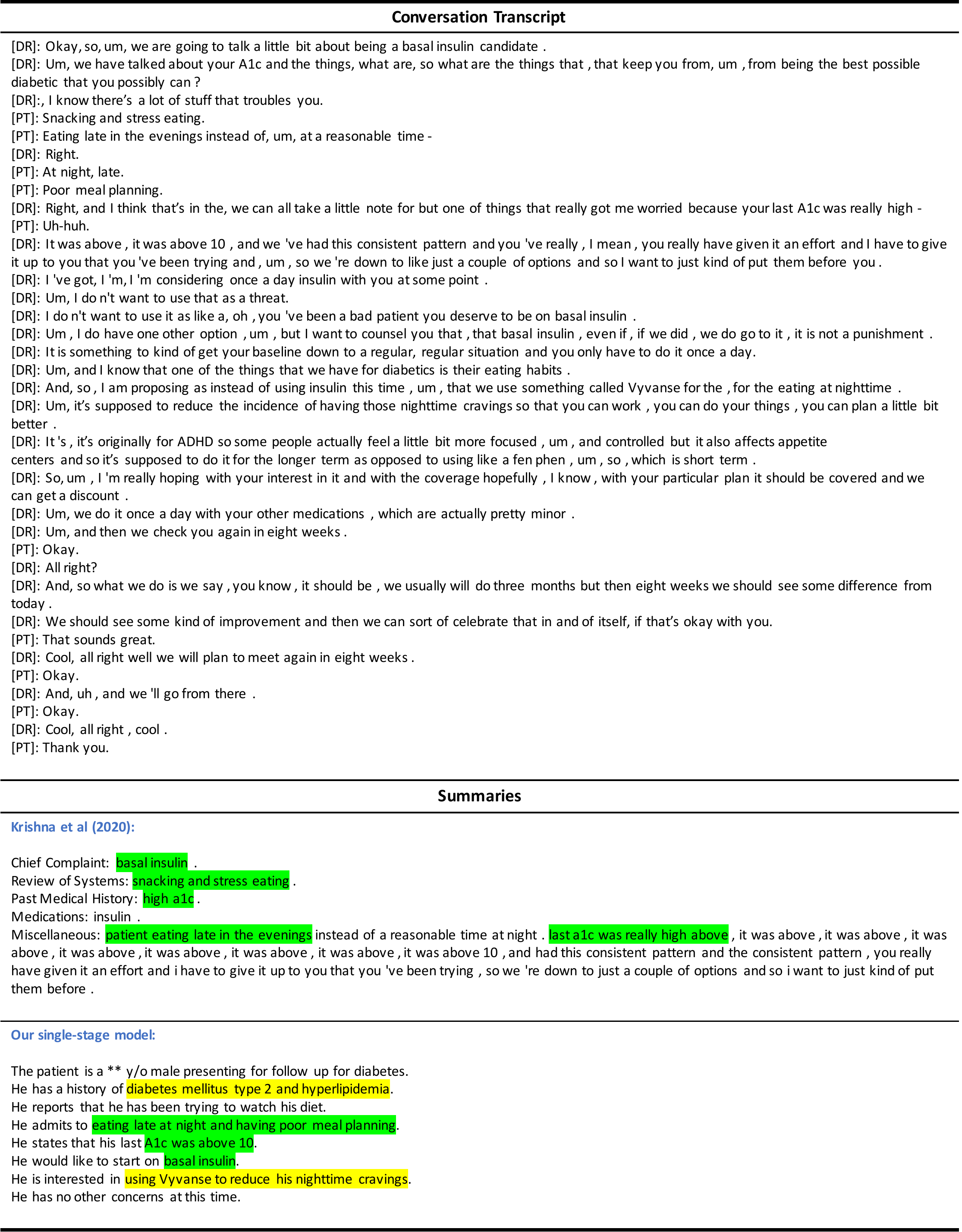}
    \caption{Inference on example conversation in \cite{krishna2020generating}. Text with \textcolor{green}{green} highlighting shows medical findings common in both summaries; text with \textcolor{yellow}{yellow} highlighting shows findings unique to our single-stage model generation that are supported by the conversation. We choose to omit the Assessment\&Plan section from the original paper as HPI and Assessment\&Plan have little overlap in contents in a medical report.}
    \label{fig:abridge_ai}
\end{figure*}

\begin{figure*}[tbhp]
    \centering
    \includegraphics[width=1.0\textwidth]{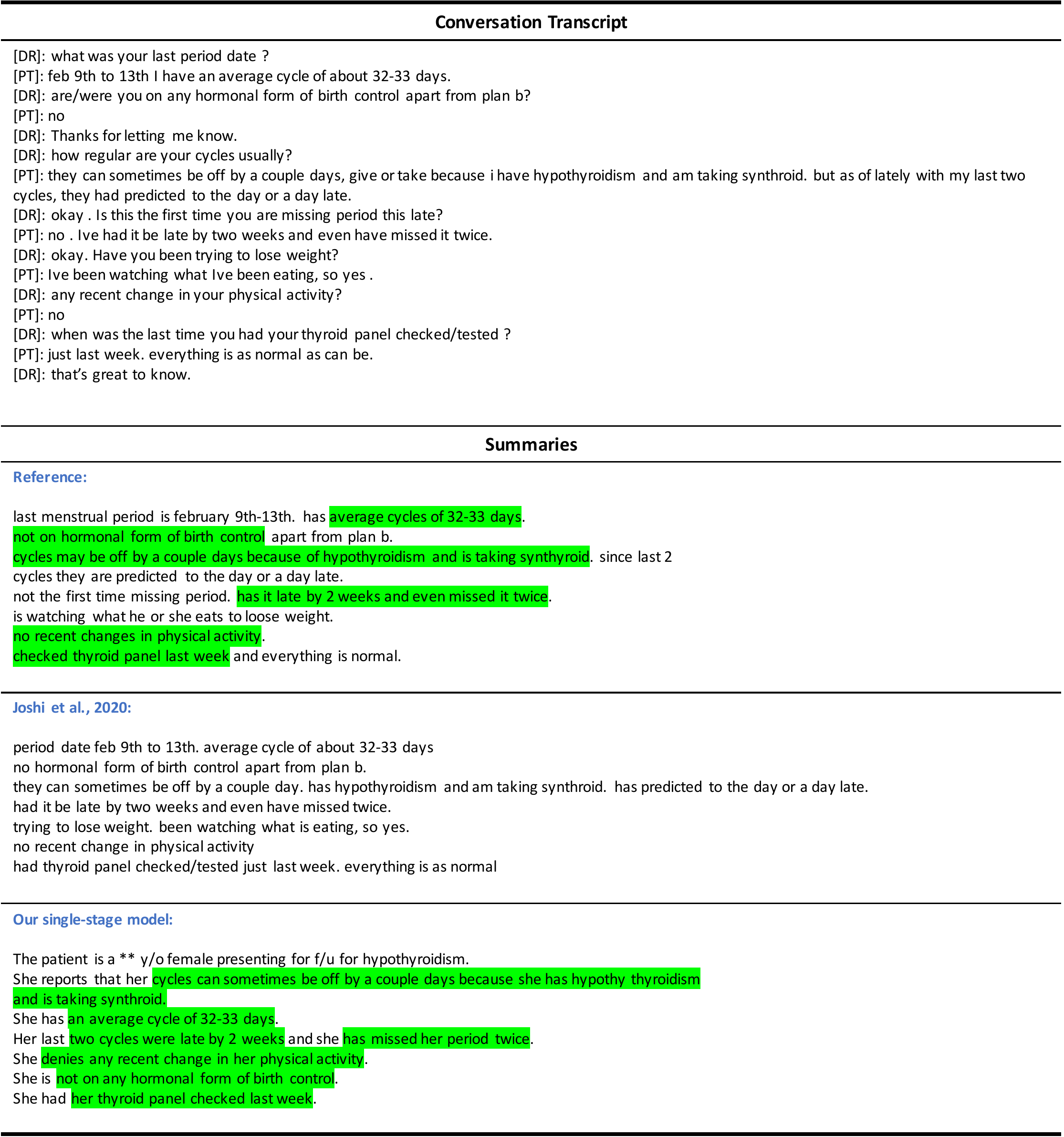}
    \caption{Inference on example conversation from \cite{joshi2020dr}. Text with \textcolor{green}{green} highlighting shows medical findings common in both summaries.}
    \label{fig:dr_summ}
\end{figure*}

Figure~\ref{fig:abridge_ai}-\ref{fig:dr_summ} display summaries generated on example conversations from \cite{krishna2020generating} and \cite{joshi2020dr}.

\subsection{Additional Evaluation Results}
\label{ssec:addi_results}
\paragraph{Hyperparameter tuning on header length} Table~\ref{tbl:rouge_ablation} shows hyperparameter tuning on the percentage of header utterances retained in all conversation chunks in the multistage (Chunking) method. All percentages are measured in unit of words, i.e., for a conversation chunk of 512 words, 25\% header means the header text spans 128 words, rounded up to the end of a turn in the original conversation. 128-word header is the setting used in this paper with the best ROUGE scores and least amount of inputs truncated in the second stage fine-tuning.

\paragraph{Development set performance} Table~\ref{tbl:test_results} shows evaluation results on the development set. Most metrics are on par or slightly worse than those obtained on the test set. Although slight overfitting was observed during model fine-tuning, the comparable model performance on both development and test set indicates that reasonable performance on unseen medical conversations of similar specialty can be expected.

\paragraph{Gender mismatch.} Roughly 30\% of the model generated summaries predict the wrong patient gender. We do not penalize such a mistake in human evaluation (Section~\ref{ssec:human_evaluation}) because (a) inferring gender is not always possible solely from the conversation transcript, nor is it necessary as this information is easily attainable; (b) the model does a good job of picking up gender pronouns if they are present in the input, but this can lead to mistakes when the gender is referring to a person other than the patient; (c) correcting gender mismatch is straightforward: we experiment with adding one sentence with the correct patient gender, \textit{The patient is a female/male}, to all model inputs and the resulting summaries predict the patient gender in $100\%$ of the observed examples.
\begin{table*}[tbhp]
\centering
\begin{tabular}{c|c|ccc}
\specialrule{0.1em}{0.1em}{0em}
& Truncated (\%) & ROUGE-1 F1      & ROUGE-2 F1      & ROUGE-l F1      \\ \hline
0\% header & 17.5 (3.7)  & 0.2893 (0.4144) & 0.0934 (0.1613) & 0.2971 (0.3930) \\
25\% header & \textbf{9.5 (1.7)}           & \textbf{0.3227 (0.4578)} & \textbf{0.1144 (0.1991)} & \textbf{0.3302 (0.4442)} \\
50\% header & 34.4 (17.6)           & 0.2921 (0.4252) & 0.0942 (0.1770) & 0.3000 (0.4100) \\
75\% header & 45.5 (30.6)           & 0.2955 (0.4212) & 0.0934 (0.1678) & 0.3023 (0.4064) \\
\specialrule{0.1em}{0em}{0.1em}
\end{tabular}
\caption{Dependence of ROUGE scores on the amount of header utterances used in Multistage (Chunking) method. Second column shows the percentage of inputs $>1024$ tokens (values in parentheses are for inputs $> 2048$ tokens) to the second stage fine-tuning. Evaluation done on dev set.}
\label{tbl:rouge_ablation}
\end{table*}

\begin{table*}[tbhp]
\resizebox{\textwidth}{!}{%
\begin{tabular}{c|ccc|ccc|ccc}
\specialrule{0.1em}{0em}{0.1em}
                      & \multicolumn{3}{c|}{ROUGE}                                                     & \multicolumn{3}{c|}{quickUMLS}                      & \multicolumn{3}{c}{rule-based}                    \\ \hline
                      & ROUGE-1 F1               & ROUGE-2 F1               & ROUGE-L F1               & F1              & Precision       & Recall          & F1              & Precision       & Recall          \\ \hline
single stage          & 0.3029 (0.4364)          & 0.1047 (0.1841)          & 0.3191 (0.4285)          & 0.3540          & 0.4229          & 0.3430          & \textbf{0.4014} & 0.5239          & \textbf{0.5097} \\
multistage (Chunking) & \textbf{0.3227 (0.4578)} & \textbf{0.1144 (0.1991)} & \textbf{0.3302 (0.4442)} & \textbf{0.3922} & \textbf{0.4764} & \textbf{0.4076} & 0.3829          & \textbf{0.5350} & 0.4877          \\
multistage (SentBERT) & 0.2997 (0.4329)          & 0.0997 (0.1691)          & 0.3098 (0.4127)          & 0.3665          & 0.4334          & 0.3646          & 0.3580          & 0.4731          & 0.4732          \\
\specialrule{0.1em}{0em}{0.1em}
\end{tabular} %
}
\caption{BART fine-tuning results on dev set.}
\label{tbl:test_results}
\end{table*}

\end{document}